\pdfoutput=1
\documentclass[11pt]{article}

\usepackage[final]{acl}

\usepackage{times}
\usepackage{latexsym}

\usepackage[T1]{fontenc}
\usepackage[utf8]{inputenc}

\usepackage{microtype}

\usepackage{inconsolata}

\usepackage{longtable}
\usepackage{xcolor}
\usepackage{colortbl}
\usepackage{multicol}
\usepackage{booktabs}
\usepackage{makecell}
\usepackage{amsmath, amssymb}
\usepackage{multirow}
\usepackage{mathtools}
\usepackage[inline]{enumitem}
\usepackage{subcaption}
\usepackage{float}
\usepackage{graphicx}
\usepackage{caption} 
\usepackage{upgreek}
\usepackage{seqsplit}
\usepackage{color, soul}
\usepackage{arydshln}
\usepackage{fix-cm}

\usepackage{amsmath,amsfonts,bm}

\def\eqref#1{equation~\ref{#1}}
\def\1{\bm{1}}

\def\vd{{\bm{d}}}

\def\vm{{\bm{m}}}

\def\vo{{\bm{o}}}
\def\vp{{\bm{p}}}

\def\vx{{\bm{x}}}
\def\vy{{\bm{y}}}

\DeclareMathAlphabet{\mathsfit}{\encodingdefault}{\sfdefault}{m}{sl}
\SetMathAlphabet{\mathsfit}{bold}{\encodingdefault}{\sfdefault}{bx}{n}

\DeclareMathOperator*{\argmax}{arg\,max}

\usepackage{caption}
\usepackage{subcaption}
\title{ResearchAgent: Iterative Research Idea Generation \\ over Scientific Literature with Large Language Models}

\author{
    Jinheon Baek$^1$ \; 
    Sujay Kumar Jauhar$^2$ \; 
    Silviu Cucerzan$^2$ \; 
    Sung Ju Hwang$^{1,3}$ \\
    KAIST$^1$ \;\; Microsoft Research$^2$ \;\; DeepAuto.ai$^3$ \\
    \texttt{\{jinheon.baek, sungju.hwang\}@kaist.ac.kr} \; \texttt{\{sjauhar, silviu\}@microsoft.com}
}

\begin{document}
\maketitle

\begin{abstract}

The pace of scientific research, vital for improving human life, is complex, slow, and needs specialized expertise. Meanwhile, novel, impactful research often stems from both a deep understanding of prior work, and a cross-pollination of ideas across domains and fields. To enhance the productivity of researchers, we propose ResearchAgent, which leverages the encyclopedic knowledge and linguistic reasoning capabilities of Large Language Models (LLMs) to assist them in their work. This system automatically defines novel problems, proposes methods and designs experiments, while iteratively refining them based on the feedback from collaborative LLM-powered reviewing agents. Specifically, starting with a core scientific paper, ResearchAgent is augmented not only with relevant publications by connecting information over an academic graph but also entities retrieved from a knowledge store derived from shared underlying concepts mined across numerous papers. Then, mimicking a scientific approach to improving ideas with peer discussions, we leverage multiple LLM-based ReviewingAgents that provide reviews and feedback via iterative revision processes. These reviewing agents are instantiated with human preference-aligned LLMs whose criteria for evaluation are elicited from actual human judgments via LLM prompting. We experimentally validate our ResearchAgent on scientific publications across multiple disciplines, showing its effectiveness in generating novel, clear, and valid ideas based on both human and model-based evaluation results. Our initial foray into AI-mediated scientific research has important implications for the development of future systems aimed at supporting researchers in their ideation and operationalization of novel work\footnote{Code: \href{https://github.com/JinheonBaek/ResearchAgent}{https://github.com/JinheonBaek/ResearchAgent}.}. 

\end{abstract}
\section{Introduction}

Scientific research plays a crucial role in driving innovation, advancing knowledge, solving problems, expanding our understanding of the world, and ultimately improving the lives of people in tangible ways. This process usually consists of two key components: the formulation of new research ideas and the validation of these ideas through well-crafted experiments, which are typically conducted by human researchers~\cite{scientificdiscovery/1, scientificdiscovery/2, Airesearchagent}. However, this is a slow, effort-intensive process, which requires reading and synthesizing overwhelming amounts of knowledge over the vast corpus of rapidly growing scientific literature to formulate research ideas, as well as design and perform experimental validations of those ideas. For example, the number of academic papers published per year is more than 7 million~\cite {paper}. Similarly, the process of testing a new pharmaceutical drug requires deep expertise, and is massively expensive and labor-intensive, often taking several years~\cite{drugdiscovery}.

\begin{figure}[t!]
    \centering
    \includegraphics[width=0.975\columnwidth]{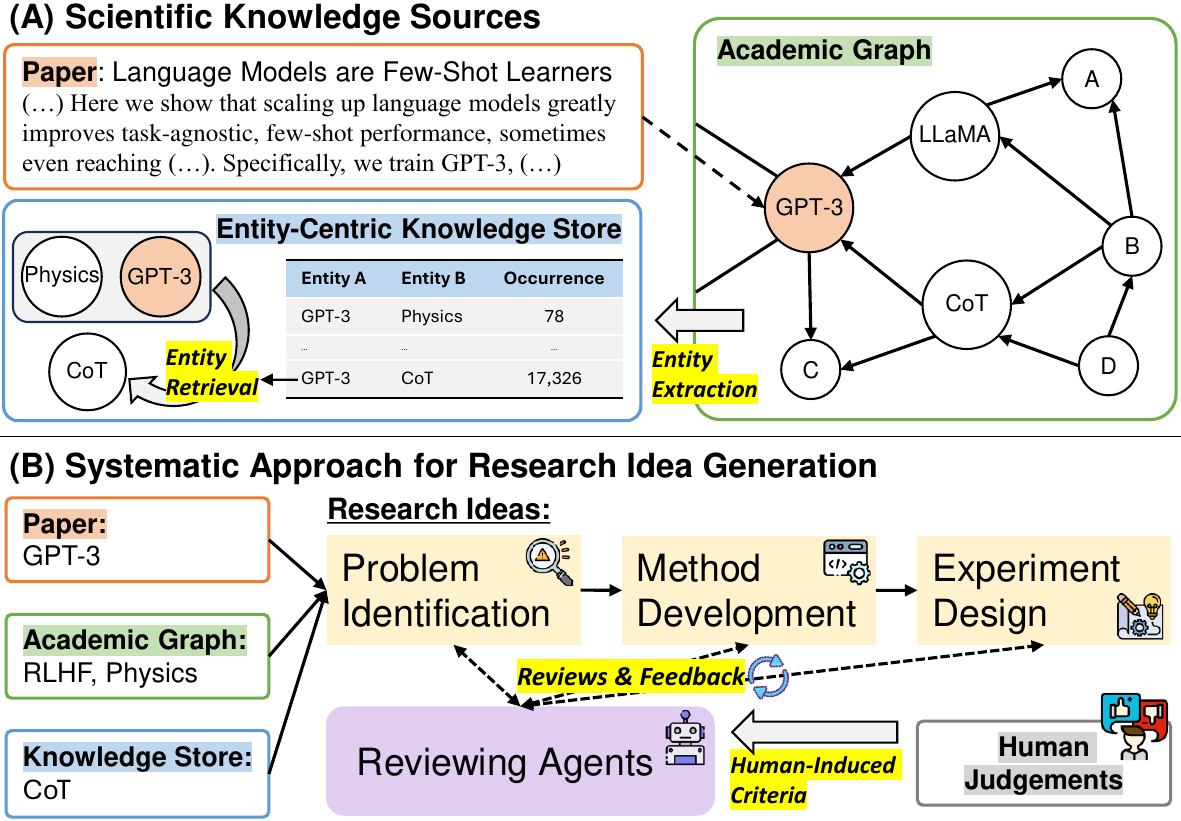}
    \vspace{-0.1in}
    \caption{(A) The scientific knowledge used for research idea generation consists of a paper, its relationships over an academic graph, and entities within a knowledge store extracted from numerous papers. (B) Given them, the proposed research idea generation process involves problem identification, method development, and experiment design. Those are also iteratively refined by reviews and feedback from reviewing agents, aligned with criteria induced from human judgements.}
    \label{fig:concept}
    \vspace{-0.125in}
\end{figure}

In the meantime, recent Large Language Models (LLMs)~\cite{Llama2, GPT-4, Gemini} have shown impressive capabilities in processing and generating text with remarkable accuracy, even outperforming human experts across diverse specialized domains including math, physics, history, law, medicine, and ethics. They are able to process and analyze large volumes of data at speeds and scales far exceeding human capabilities, have internalized large swaths of human knowledge from being trained on virtually the entire web, and can identify patterns, trends, and correlations that may not be immediately apparent to human researchers (such as the usage of quantum mechanics in medical imaging or applying psychological insights in AI). This renders them ideally poised to become foundational tools to accelerate the two phases of the scientific research process: ideation of novel research opportunities, and scientific validation of those research hypotheses.

A few recent papers in the domain of LLM-augmented scientific discovery have focused on the second phase. Specifically, they attempt~\cite{Airesearchagent, LLM/discovery/1, LLM/discovery/2} to mainly accelerate the experimental validation process, by writing code for machine-learning models, facilitating the exploration of chemical spaces, or advancing the simulation of molecular dynamics.
Thus, in this paper, we leverage LLMs in the first phase of scientific research -- specifically idea generation, whose key focus is conceptualizing novel research questions, methodologies, and experiments. To the best of our knowledge, our work is the first to leverage and evaluate the capabilities of LLMs to act as mediators in scientific idea generation in an open-ended setting.

Given our goal to build an LLM-powered ResearchAgent, we draw inspiration from how human researchers position themselves to come up with novel research ideas. We draw distinctions between three key components of their workflow: a broad and deep understanding of related scientific literature, an encyclopedic view of concepts and how they relate to one another both within and across domains, and a community of colleagues on which to rely for feedback and constructive criticism.

We model each of these three aspects in our ResearchAgent. Specifically, in order to imbibe related work, the system begins with a core scientific paper and then explores a range of related papers through references and citation relationships. Further, to develop an encyclopedic view of related concepts, we build and then augment ResearchAgent with an entity-centric knowledge store derived from co-occurrences of key concepts in the scientific literature. This repository is aimed at capturing novel underlying relationships within and across domains, thereby increasing the chances of a cross-pollination of ideas~\cite{cross-pollination}. Finally, to simulate robust feedback mechanisms, we instantiate a number of LLM-powered ReviewingAgents that help the ResearchAgent to iterate on research idea generation with constructive critiques. Crucially, these ReviewingAgents are prompted with evaluation criteria that are induced from real researchers' judgements, thus aligning them with actual scientific preferential standards. An illustration of our system is provided in Figure~\ref{fig:concept}.

We validate the effectiveness of ResearchAgent for research idea generation based on scientific literature across multiple disciplines. Then, on a battery of tests conducted with both human- and model-based evaluations, we demonstrate that ResearchAgent outperforms strong LLM-powered baselines by large margins, generating more clear, relevant, and significant ideas that are especially novel. Furthermore, analyses show the efficacy of our comprehensive approach to modeling ResearchAgent: the entity-centric knowledge store and the iterative idea refinement steps help the system generate meaningfully better ideas compared with an instantiation that is purely based on prior related work.

These findings highlight the immense potential of AI-mediated research assistants like ResearchAgent to enhance the ideation process in scientific research. In practice, it can support researchers by identifying knowledge gaps, proposing novel problem statements, and suggesting potential methodologies early in the research process. Also, it can assist in designing experiments and streamline the writing and refinement of research papers by generating drafts and offering feedback on how to effectively frame contributions and cite relevant work.

\section{Related Work}
\label{sec:relatedwork}

\paragraph{Large Language Models}
LLMs have shown impressive performances across various tasks~\cite{GPT-4, Gemini}, including scientific fields such as mathematics, physics, medicine, and computer science~\cite{author/discovery, MathematicalLLM, LLM/discovery/2, Airesearchagent, idea/co-creation}. For instance, GPT-4 can understand DNA sequences, design biomolecules, predict molecular behavior, and solve PDE problems~\cite{LLM/discovery/1}. However, LLMs have mainly been used for accelerating the experimental validation of already identified research ideas, but not for identifying new problems. 

\paragraph{Hypothesis Generation}
The principle of hypothesis generation is based on literature-based discovery~\cite{LBD}, which aims to discover relationships between concepts~\cite{Henry2017LiteratureBD}. For instance, these concepts could be a specific disease and a compound not yet considered as a treatment for it. Early works on automatic hypothesis generation first build a corpus of discrete concepts, and then identify their relationships with machine learning approaches, e.g., using similarities between word (concept) vectors~\cite{hypothesis/wordvector} or applying link prediction methods over a graph (where concepts are nodes)~\cite{hypothesis/graph/1, hypothesis/graph/2}. Recent approaches are further powered by LLMs~\cite{CLBD, zeroshot/hypothesis, tomato}, leveraging their prior knowledge about scientific disciplines. Yet, all these approaches perform idea generation in a localized manner and are designed to identify potential relationships between two variables or generate sentence-level connections, which may be sub-optimal to capture the complexity and multifaceted nature of real-world problems (e.g., urban planning involves numerous interacting variables). Meanwhile, we do not artificially restrict the target research idea to be a predictive single concept or simple binary link, instead allowing the model to generate ideas in a more open-ended fashion.

We note that there has been a recent surge of interest in exploring scientific idea generation: from \citet{li2024chainideasrevolutionizingresearch} that focus on evaluating whether LLMs can generate research ideas that are better than human ideas, to \citet{lu2024aiscientistfullyautomated} that aim to automatically generate full research papers (including idea development, code writing, and experiment execution), to \citet{li2024chainideasrevolutionizingresearch} that enhance the idea generation process by organizing a sequential chain of literature, all of which acknowledge and build upon insights from a prior version of our paper.

\paragraph{Knowledge-Augmented LLMs}
The approach to augment LLMs with external knowledge makes them more accurate and relevant to target contexts. Much prior work aims at improving the factuality of LLM responses to queries by retrieving the relevant documents and injecting them into the LLM input~\cite{InternetaugmentedLM, InContextRetrieval, REPLUG}. In addition, given that entities or facts are atomic units for representing knowledge, recent studies augment LLMs with them~\cite{KAPING, RetrieveRewriteAnswer}. In contrast to these efforts, which use knowledge units piecemeal, we instead jointly leverage accumulated knowledge over massive troves of scientific papers. Also, \citet{K-LaMP} proposes to use entities for query suggestion, which -- while similar -- has the entirely different objective of narrowing the focus of LLMs to entities already present in their context. Instead, our approach retrieves and integrates entities outside the given context, enabling LLMs to explore other concepts for research idea generation.

\paragraph{Iterative Refinements with LLMs}
Similar to humans, LLMs do not always generate optimal outputs on their first attempt. To tackle this, drawing inspiration from humans who can iteratively refine their thoughts based on critiques from themselves and their peers, many recent studies have investigated the potential of LLMs to correct and refine their outputs, demonstrating that they indeed possess those capabilities~\cite{self-refine/0, self-refine/1, self-refine/2, self-refine/3, CLBD, zeroshot/hypothesis, tomato}. Based on their findings, we extend this paradigm (and further test their capability) to our novel scenario of research idea generation.

\section{Method}
We present ResearchAgent, a system that automatically proposes research ideas with LLMs.

\subsection{LLM-Powered Research Idea Generation}
\label{sec:research-idea-gen}

We begin by formally introducing the new problem of research idea generation, followed by an explanation of how LLMs are utilized to tackle it.

\paragraph{Research Idea Generation}
The goal of the research idea generation task is to formulate new and valid research ideas, to enhance the overall efficiency of the first phase of scientific discovery. While we acknowledge that the real process by which humans conduct research is varied and complex to an extent well beyond the scope of this scientific study, we attempt to model simulacra in three systematic steps that would likely be maximally beneficial to a researcher seeking assistance from an AI system. These are namely, identifying novel research ideas, proposing methods to validate these ideas, and designing experiments to measure the success of these methods in relation to the ideas.

To accomplish the aforementioned steps, we utilize the existing literature (such as academic publications) as a primary source, which provides insights about existing knowledge along with gaps and unanswered questions\footnote{We focus on the existing literature-based idea generation by following the paradigm that a \textit{new idea} is more often than not just a new combination of old elements~\cite{young2003technique}.}. Formally, let $\mathcal{L}$ be the literature, and $\vo$ be the ideas that consist of the problem $\vp$, method $\vm$, and experiment design $\vd$, as follows: $\vo = [\vp, \vm, \vd]$ where each item consists of a sequence of tokens. Then, the idea generation model $f$ can be represented as follows: $\vo = f(\mathcal{L})$, which is further decomposed into three submodular steps: $\vp = f(\mathcal{L})$ for identifying problems, $\vm = f(\vp, \mathcal{L})$ for developing methods, and $\vd = f(\vp, \vm, \mathcal{L})$ for designing experiments. We operationalize $f$ with LLMs, leveraging their capability to understand and generate academic text.

\paragraph{Large Language Models}
Before describing the LLM in the context of our problem setup, let us first provide its general definition, which takes an input sequence of tokens $\vx$ and generates an output sequence of tokens $\vy$, as follows: $\vy = \texttt{LLM}_{\theta}(\mathcal{T}(\vx))$. Here, the model parameters $\theta$ are typically fixed after training, due to the high costs of further fine-tuning. In addition, the prompt template $\mathcal{T}$ serves as a structured format that outlines the context (including the task descriptions and instructions) to direct the model in generating the desired outputs.

\subsection{Knowledge-Augmented LLMs for~Research~Idea~Generation}
We now turn to our primary focus of automatically generating research ideas with LLMs. Recall that we aim to produce a complete idea consisting of the problem, method, and experiment design ($\vo = [\vp, \vm, \vd]$), while using the existing literature $\mathcal{L}$ as a primary source of information. We operationalize this with LLMs by instantiating the aforementioned research idea generation function $f$ with $\texttt{LLM}$ coupled with the task-specific template. Formally, $\vp = \texttt{LLM}(\mathcal{T}_p(\mathcal{L}))$ indicates the problem identification step, followed by $\vm = \texttt{LLM}(\mathcal{T}_m(\vp, \mathcal{L}))$ for method development and $\vd = \texttt{LLM}(\mathcal{T}_e(\vp, \vm, \mathcal{L}))$ for experiment design, which constitutes the full idea: $\vo = [\vp, \vm, \vd]$. 

Following this general formulation, the important question to answer is how the body of scientific literature is leveraged for actually generating research ideas with LLMs. 
Here, we outline three key desiderata that contribute to the success of human researchers ideating novel research ideas: a broad and deep understanding of related work, an encyclopedic perspective on the interconnectedness of concepts within and across scientific domains, and a community of peers who help iteratively improve ideas through constructive critiques. We describe our operationalization of these three desiderata using the prior literature and LLMs in what follows.

\paragraph{Citation Graph-based Literature Survey}
Due to the constraints on their input lengths and their reasoning abilities, particularly over very long contexts~\cite{LostITM}, it is not possible to incorporate all the existing publications from the literature $\mathcal{L}$ into the LLM input. Instead, we need to find a meaningful subset relevant to the problem at hand. To achieve this, we mirror the process followed by human researchers, who expand their knowledge of a paper by perusing other papers that either cite or are cited by it. Concretely, for the $\texttt{LLM}$, we initiate its literature review process by providing a core paper $l_0$ from $\mathcal{L}$ and then selectively incorporating subsequent papers $\left\{ l_1, ..., l_n \right\}$ that are directly connected based on a citation graph. This procedure makes the $\texttt{LLM}$ input for idea generation more manageable and coherent. In addition, we operationalize the selection process of the core paper and its relevant citations with two design choices: 1) the core paper is selected based on its citation count (e.g., exceeding 100 over 3 months) typically indicating high impact; 2) its relevant papers (which may be potentially numerous) are further narrow-downed based on their similarities of abstracts with the core paper, ensuring a more focused and relevant set of related work.

However, despite the simplicity and intuitiveness of this idea generation approach, there exists one major limitation. This approach relies exclusively on a set of given papers (the core paper and its citations); however, since scientific knowledge is not confined to specific studies but rather accumulates across a wide range of publications (across various fields), we should ideally harness this extensive, interconnected, and relevant scientific knowledge in our method for research idea generation.

\paragraph{Entity-Centric Knowledge Augmentation}
In order to model an encyclopedic view of interconnected concepts, we must effectively design a framework to extract, store and effectively leverage the vast amount of knowledge in scientific literature $\mathcal{L}$. In this work, we view entities as the atomic units of knowledge, which allows for ease of representation and accumulation over papers in a unified manner across different disciplines. For example, we can easily extract the term ``database'' whenever it appears in any paper, using existing off-the-shelf entity linking methods and then aggregate their linked occurrences into a knowledge store. Then, if the term ``database'' is prevalent within the realm of medical science but less so in hematology (which is a subdomain of medical science), the constructed knowledge store can capture the affinity between those two domains based on overlapping entities. This representational paradigm can then be used to suggest the term ``database'' when formulating the ideas about hematology. In other words, this approach enables providing novel and interdisciplinary insights by leveraging the interconnectedness of entities across various fields.

Formally, we design the knowledge store as a two-dimensional matrix $\mathcal{K} \in \mathcal{R}^{m \times m}$ where $m$ is the total number of unique entities identified and $\mathcal{K}$ is implemented in a sparse format. This knowledge store is constructed by extracting entities over all the available scientific articles in literature $\mathcal{L}$\footnote{As extracting entities on all articles is computationally infeasible, we target papers appearing after May 01, 2023.}, which not only counts the co-occurrences between entity pairs within individual papers but also quantifies the count for each entity. Our approach is versatile, thus, we can use any entity linker; in this paper we use one developed by~\citet{blink}. This off-the-shelf system proves capable of extracting key scientific entities (which is demonstrated in Table~\ref{tab:examples}) despite its lack of customized training for the scientific domain. Specifically, this linker tags and canonicalizes entities in a paper $l$ from $\mathcal{L}$, formalized as follows: $\mathcal{E}_l = \texttt{EL}(l)$ where $\mathcal{E}_l$ denotes a multiset of entities (allowing for repetitions) appearing in $l$\footnote{Due to the extensive length of scientific publications, the target of entity extraction is restricted to titles and abstracts.}. Upon extracting entities $\mathcal{E}$, to store them into the knowledge store $\mathcal{K}$, we consider all possible pairs of $\mathcal{E}$ represented as follows: $\left\{ e_i, e_j \right\}_{(i, j) \in \mathcal{C}(|\mathcal{E}|, 2)}$ where $e \in \mathcal{E}$.

Given this knowledge store $\mathcal{K}$, our next goal is to enhance the previous vanilla research idea generation process implemented based on a group of interconnected papers, denoted as follows: $\vo = \texttt{LLM}(\mathcal{T}(\left\{ l_0, l_1, ..., l_n \right\}))$. We do this by augmenting the $\texttt{LLM}$ with the relevant entities from $\mathcal{K}$, which expand the context that LLMs consume with additional knowledge. Formally, let us define entities extracted from the group of interconnected papers, as follows: $\mathcal{E}_{\left\{ l_0, ..., l_n \right\}} = \bigcup_{i=0}^{n} \texttt{EL}(l_i)$. Then, the probabilistic form of retrieving the top-$k$ relevant external entities can be represented as follows: 
\begin{equation}
\fontsize{8pt}{8pt}\selectfont
    \texttt{Ret}(\left\{ l_0, ..., l_n \right\}; \mathcal{K}) = \argmax_{I \subset [m]: |I| = k } \prod {P(e_i | \mathcal{E}_{\left\{ l_0, ..., l_n \right\}})},
    \label{eq:retrieval}
\fontsize{8pt}{8pt}\selectfont
\end{equation}
where $ [m] = \{1, ..., m\} $ and $e_i \notin \mathcal{E}_{\left\{ l_0, ..., l_n \right\}}$. Also, for simplicity, by applying Bayes' rule and assuming that entities are independent, the retrieval operation (Equation~\ref{eq:retrieval}) can be approximated as follows:
\begin{equation}
\fontsize{8.5pt}{8.5pt}\selectfont
    \argmax_{I \subset [m]: |I| = k } \prod ({\prod_{e_j \in \mathcal{E}_{\left\{ l_0, ..., l_n \right\}}} P(e_j | e_i)}) \times P(e_i),
    \label{eq:retrieval:detail}
\fontsize{8.5pt}{8.5pt}\selectfont
\end{equation}
where $P(e_j | e_i)$ and $P(e_i)$ can be derived from values in the two-dimensional matrix $\mathcal{K}$, suitably normalized. We note that the formulation in Equation~\ref{eq:retrieval:detail} is only one instance of operationalizing retrieval; this could be replaced with other retrieval strategies -- for example, embedding-based retrieval (discussions and results are provided in Appendix~\ref{appendix:retrieval}). Hereafter, the instantiation of research proposal generation augmented with relevant entity-centric knowledge is formalized as follows: $\vo = \texttt{LLM} (\mathcal{T}(\left\{ l_0, ..., l_n \right\}, \texttt{Ret}(\left\{ l_0, ..., l_n \right\}; \mathcal{K})))$\footnote{There may be additional knowledge sources (beyond the existing literature and entities) for research idea generation, and we leave exploring them as future work.}. We call this knowledge-augmented LLM-powered idea generation approach \emph{ResearchAgent}, and provide the templates to instantiate it in Tables~\ref{tab:prompt:problem}, \ref{tab:prompt:method}, and \ref{tab:prompt:experiment}.

\paragraph{Iterative Research Idea Refinements}
We note that attempting to write a full research idea in one go may not be an effective strategy. Humans write drafts that are continually improved based on multiple rounds of reviews and feedback. Therefore, we lastly model a community of peers for iterative idea improvement by introducing a set of LLM-powered reviewing agents (called \emph{ReviewingAgents}), which provide the ResearchAgent with reviews and feedback according to various criteria for improvement.

Specifically, similar to our approach to instantiate ResearchAgent with an LLM ($\texttt{LLM}$) and template ($\mathcal{T}$), ReviewingAgents are instantiated similarly but with different templates (See Tables~\ref{tab:prompt:problem:valid}, \ref{tab:prompt:method:valid}, and \ref{tab:prompt:experiment:valid}). Then, with ReviewingAgents, each of the generated research ideas (problem, method, and experiment design) is separately evaluated according to its own specific five criteria\footnote{We select the top five criteria which we consider as the most important, and leave exploring others as future work.}, which are provided in labels of Figure~\ref{fig:main} and detailed in Table~\ref{tab:criteria}. Based on the reviews and feedback from ReviewingAgents, the ResearchAgent iteratively updates and refines its generation of research ideas.

Despite the proficiency of LLMs in the evaluation of machine-generated texts~\cite{LLM/Judge/1, LLM/Judge/2}, their judgments on research ideas may not be aligned with the judgments of real human researchers. However, there are no ground truth reference judgments available, and collecting them to align LLMs is expensive and often infeasible. Ideally, the judgments made by LLMs should be similar to the ones made by humans, and we aim to ensure this by automatically generating human preference-aligned evaluation criteria (used for automatic evaluations) with a few human annotations. Specifically, to obtain these human-aligned evaluation criteria, we first collect 10 pairs of research ideas and their associated scores for every evaluation criterion on a 5-point Likert scale, annotated by human researchers having at least 3 papers. After that, we prompt the LLM with these human-annotated pairs to induce detailed descriptions for evaluation criteria~\cite{lin2024interpretable} (See Tables~\ref{tab:criteria:problem},~\ref{tab:criteria:method}, and~\ref{tab:criteria:experiment}). These criteria reflect the underlying human preferences\footnote{We additionally ask five human annotators, who evaluate research ideas, to judge the quality of the induced criteria; two of them agree strongly, while the other three agree moderately.} and are used as evaluation criteria by the ReviewingAgents.

\section{Experimental Setup}
\label{sec:setups}

\subsection{Data}
The main source to generate research ideas is the scientific literature $\mathcal{L}$, which we obtain from the Semantic Scholar Academic Graph API\footnote{\href{https://www.semanticscholar.org/product/api}{https://www.semanticscholar.org/product/api}}. From this, we select papers appearing after May 01, 2023, because LLMs that we use in our experiments are trained on data from the open web available before this point. This follows the procedure of existing literature-based hypothesis generation work~\cite{zeroshot/hypothesis}. Then, we select high-impact papers (that have more than 20 citations) as core papers, mirroring human researchers' tendency to leverage influential work, to ensure the high quality of generated ideas. The resulting data is still very large; thus, we further sample a subset of 300 papers as core papers to obtain a reasonably sized benchmark dataset. The average number of reference papers for each core paper is 87; the abstract of each paper has 2.17 entities on average. The distribution of disciplines for all papers is provided in Figure~\ref{fig:discipline}.

\subsection{Baselines and Our Model}
As we target the novel task of research idea generation involving the generation of problems, methods, and experimental designs, there are no baselines for direct comparison. Thus, we mainly compare our full ResearchAgent model against its ablated variants, outlined as follows:
\begin{enumerate}[itemsep=1.0mm, parsep=1pt, leftmargin=*]
    \item {\bf Naive ResearchAgent} -- which uses only a core paper to generate research ideas. 
    \item {\bf ResearchAgent w/o Entity Retrieval} -- which uses the core paper and its relevant references without considering entities.
    \item {\bf ResearchAgent} -- which is our full model that uses the relevant references and entities along with the core paper, to augment LLMs.
\end{enumerate}
In addition to this set of core baselines, we also compare our approach against existing hypothesis generation work from prior literature in Table~\ref{tab:hypothesis}.

\subsection{Evaluation Setup}
Given our formulation of idea generation~(Sec~\ref{sec:research-idea-gen}), there are no ground-truth answers to measure the quality of the generated ideas. Yet, exhaustively listing pairs of core papers and reference research ideas is suboptimal, since there may exist a large number of valid research ideas for each core paper, and this process requires much time, effort and expertise on the part of human researchers. Thus, we use a combination of model-based automatic evaluation and manual human evaluation to validate different models on our experimental benchmark.

\begin{figure*}[t!]
    \centering
    \begin{subfigure}[t]{0.99\linewidth}
        \centering
        \includegraphics[width=0.75\linewidth]{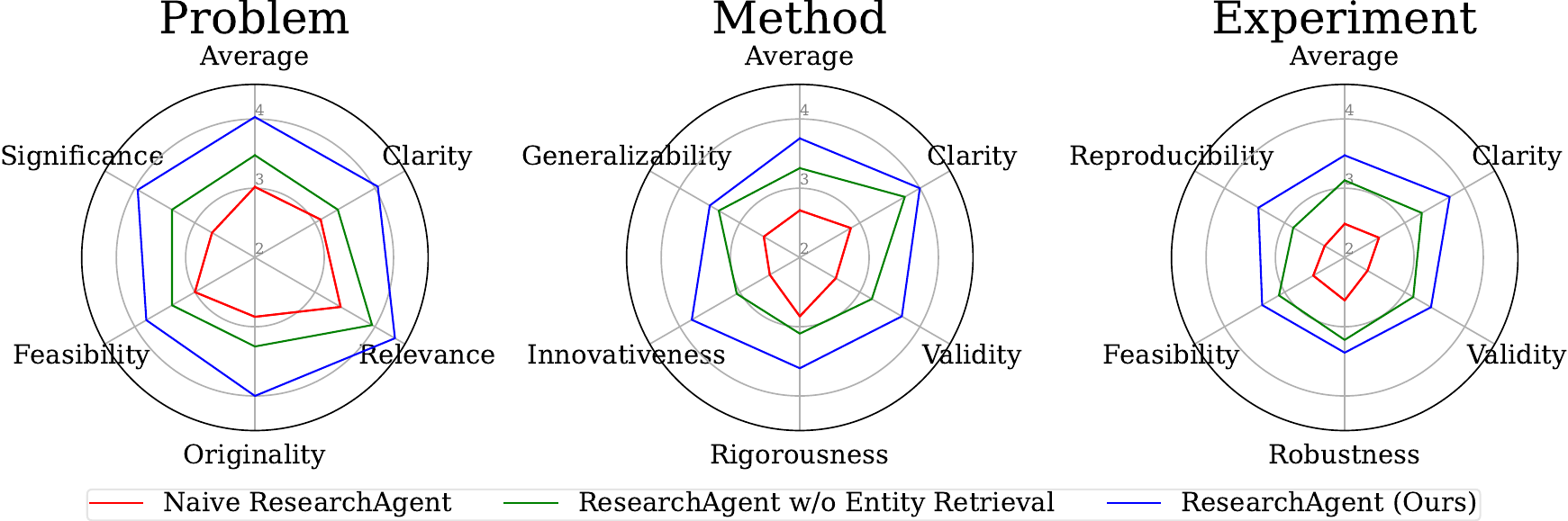}
        \caption{Human Evaluation}
    \end{subfigure}
    
    \vspace{0.1in}
    
    \begin{subfigure}[t]{0.99\linewidth}
        \centering
        \includegraphics[width=0.75\linewidth]{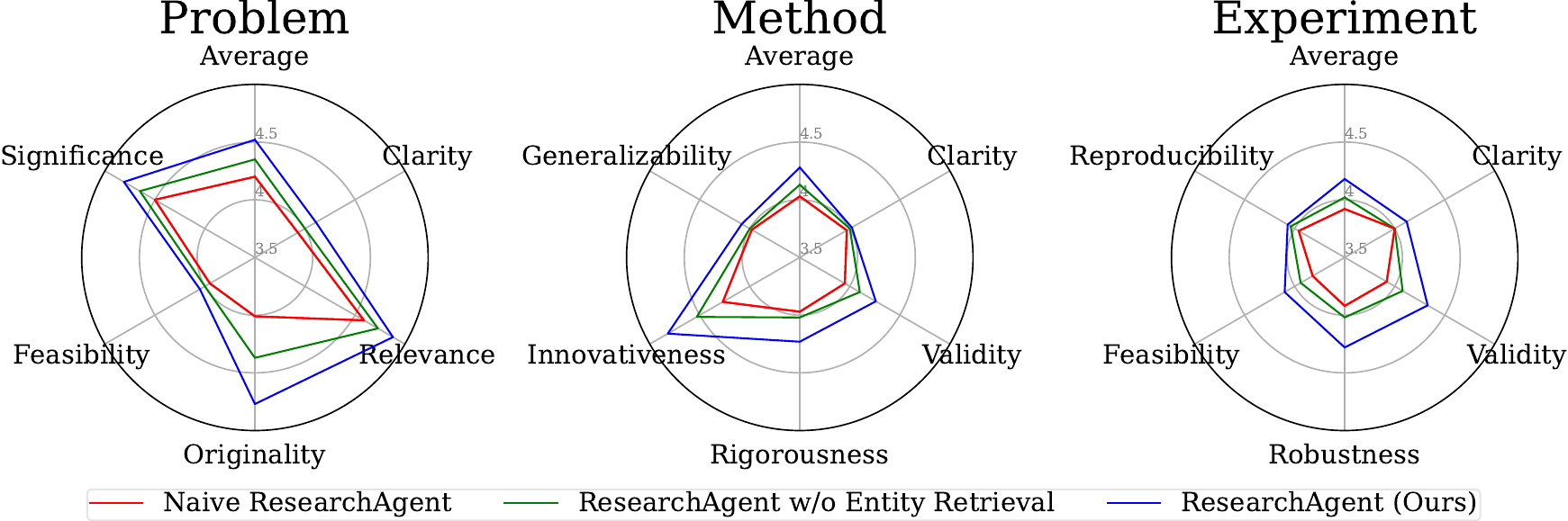}
        \caption{Model-based Evaluation}
    \end{subfigure}
    
    \vspace{-0.075in}
    \caption{Main results on our research idea generation task with human- (top) and model-based (bottom) evaluations, where we report the score of each idea (problem, method, or experiment design) based on its own five criteria and their average score.}
    \label{fig:main}
    \vspace{-0.05in}
\end{figure*}

\begin{figure}[t!]
    \centering
    \includegraphics[width=0.975\columnwidth]{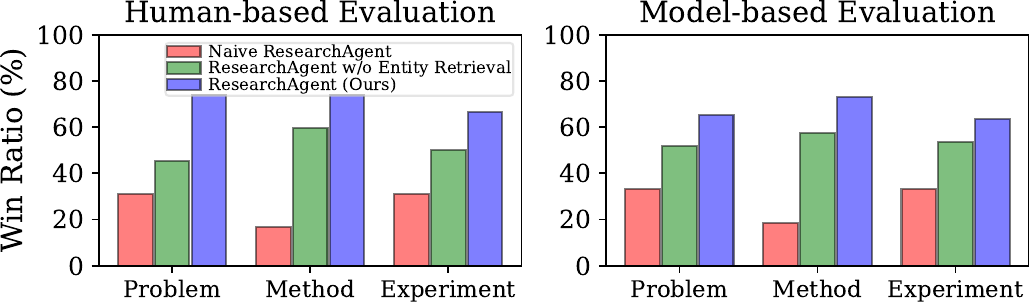}
    \vspace{-0.075in}
    \caption{Results of pairwise comparisons between ideas from two of any different approaches, where we report the win ratio.}
    \label{fig:pairwise}
    \vspace{-0.1in}
\end{figure}

\paragraph{Model-based Evaluation}
Following the recent trends in using LLMs to judge the quality of output texts (especially in the setting of reference-free evaluations)~\cite{LLM/Judge/1, LLM/Judge/2, G-Eval}, we use GPT-4 to judge the quality of research ideas. Note that each of the problem, method, and experiment design is evaluated with five different criteria (See labels of Figure~\ref{fig:main} for criteria and see Table~\ref{tab:criteria} for their detailed descriptions). We ask the LLM-based evaluation model to either rate the generated idea on a 5-point Likert scale for each criterion or perform pairwise comparisons between two ideas from different models. We provide the prompts for evaluations in Appendix~\ref{appendix:setups}.  

\paragraph{Human Evaluation}
Similar to model-based evaluations, we perform human evaluations that involve assigning a score for each criterion and conducting pairwise comparisons between two ideas. As the generated ideas are knowledge-intensive, we carefully select annotators who are well-versed in the field and provide them with ideas that are highly relevant to their field of expertise\footnote{We also experiment with human evaluation using non-domain-experts, but this proves to be suboptimal therefore, we focus on experts for reliable judgments of generated ideas.}. Specifically, we choose ten expert researchers who have authored at least three papers and ask them to judge only the ideas that are generated based on their own papers.

\subsection{Implementation Details}
We mainly use the GPT-4~\cite{GPT-4} release from Nov 06, 2023, as the basis for all models, which is, notably, reported to be trained with data up to Apr 2023 (meanwhile, the papers used for idea generation appear after May 2023). To extract entities and build the entity-centric knowledge store, we use the off-the-shelf BLINK entity linker~\cite{blink}, with papers from May 01, 2023, to Dec 31, 2023 (available from Semantic Scholar API) along with their references, which number 50,091 in total. We provide detailed prompts used to elicit responses for research idea generation in Appendix~\ref{appendix:prompts:generation}.

\begin{table}[t!]
\caption{Results of agreements between two human annotation results and between human and model evaluation results. }
\vspace{-0.075in}
\label{tab:agreement}
\small
\centering
\resizebox{0.475\textwidth}{!}{
\renewcommand{\arraystretch}{1.0}
\renewcommand{\tabcolsep}{1.0mm}
\begin{tabular}{llccc}
\toprule

\textbf{Categories} & \textbf{Metrics} & \textbf{Problem} & \textbf{Method} & \textbf{Experiment} \\

\midrule
\midrule

\multirowcell{2}[-0.0ex][l]{\textbf{Human and Human}} 

& Scoring & 0.83 & 0.76 & 0.67 \\

& Pairwise & 0.62 & 0.62 & 0.41 \\

\midrule

\multirowcell{2}[-0.0ex][l]{\textbf{Human and Model}} 

& Scoring & 0.64 & 0.58 & 0.49 \\

& Pairwise & 0.71 & 0.62 & 0.52 \\

\bottomrule

\end{tabular}
}
\vspace{-0.025in}
\end{table}

\section{Experimental Results and Analyses}
\label{sec:results}

\paragraph{Main Results}
Our main results on scoring with human and model-based evaluations are provided in Figure~\ref{fig:main}. These demonstrate that our full ResearchAgent outperforms all baselines by large margins on every metric across problems, methods, and experiment designs (constituting the complete research ideas). Particularly, the full ResearchAgent augmented with relevant entities exhibits strong gains on metrics related to creativity (such as Originality for problems and Innovativeness for methods) since entities may offer novel concepts and views that may not be observable in the group of citation-based papers alone. In addition, the results of pairwise comparisons between models with both human and model-based evaluations -- shown in Figure~\ref{fig:pairwise} -- demonstrate that the full ResearchAgent shows the highest win ratio over its baselines.  

\paragraph{Analysis on Inter-Annotator Agreements}
To validate the quality and reliability of human annotations, we measure the inter-annotator agreements, where 20\% of the generated ideas are evaluated by two human judges, and report results in Table~\ref{tab:agreement}. Specifically, for the scoring, we first rank scores from each annotator and measure Spearman's correlation coefficient~\cite{spearman} between the ranked scores of two annotators. For the pairwise comparison between two judges, we measure Cohen’s kappa coefficient~\cite{cohen}. Table~\ref{tab:agreement} shows that the inter-annotator agreement is high, confirming the reliability of our assessments about the quality of generated research ideas. Also, while agreement scores for experimental designs are slightly lower than other aspects, this does not necessarily indicate a shortcoming in the quality of experimental designs produced by ResearchAgent, as demonstrated in Figures~\ref{fig:main} and~\ref{fig:pairwise}. Instead, we view this as the inherent subjectivity and variability in how such designs are perceived and evaluated by different annotators (i.e., the nature of the variability itself makes achieving high agreement challenging).

\begin{figure}[t!]
    \centering
    \includegraphics[width=0.985\columnwidth]{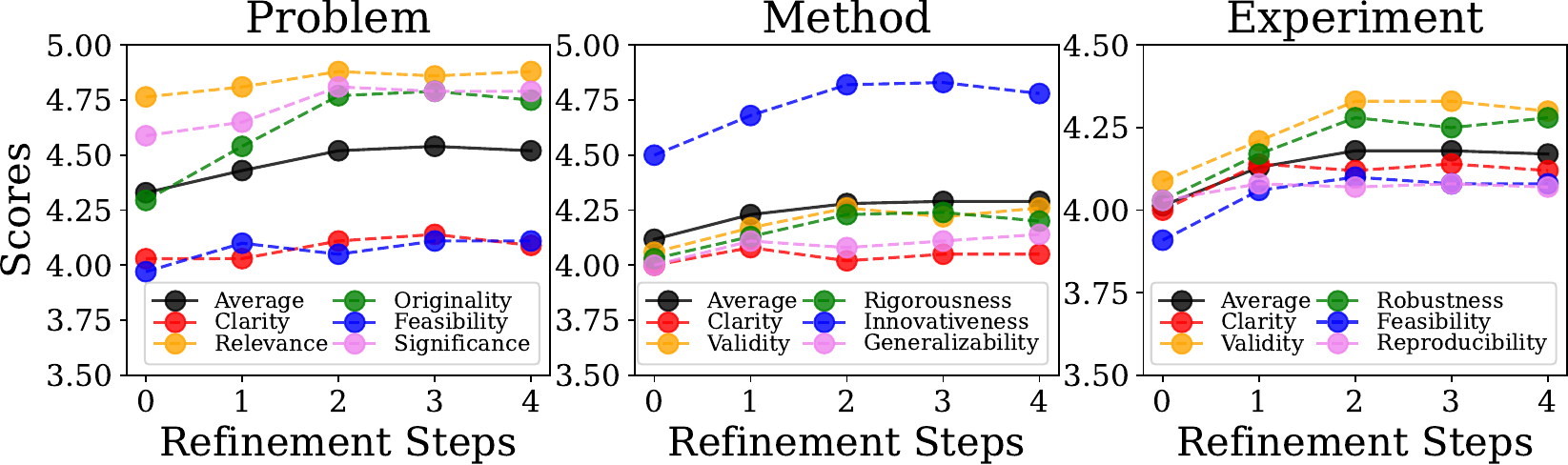}
    \vspace{-0.08in}
    \caption{Results with varying the number of refinement steps.}
    \label{fig:iteration}
    \vspace{-0.07in}
\end{figure}
\begin{table}[t!]
\caption{Results of ablation study on references and entities.}
\vspace{-0.1in}
\label{tab:ablation}
\small
\centering
\resizebox{0.475\textwidth}{!}{
\renewcommand{\arraystretch}{0.8}
\begin{tabular}{lccc}
\toprule

\textbf{Methods} & \textbf{Problem} & \textbf{Method} & \textbf{Experiment} \\

\midrule
\midrule

ResearchAgent & \textbf{4.52} & \textbf{4.28} & \textbf{4.18} \\

\noalign{\vskip 0.25ex}\cdashline{1-4}\noalign{\vskip 0.75ex}

- w/o Entities & 4.35 & 4.13 & 4.02 \\

- w/ Random Entities & 4.41 & 4.19 & 4.13 \\

\noalign{\vskip 0.25ex}\cdashline{1-4}\noalign{\vskip 0.75ex}

- w/o References & 4.26 & 4.08 & 3.97 \\

- w/ Random References & 4.35 & 4.16 & 4.02 \\

\noalign{\vskip 0.25ex}\cdashline{1-4}\noalign{\vskip 0.75ex}

- w/o Entities \& References & 4.20 & 4.03 & 3.92 \\

\bottomrule

\end{tabular}
}
\vspace{-0.1in}
\end{table}

\paragraph{Analysis on Human-Model Agreements} Similar to what we did for the aforementioned inter-annotator agreements, we measure agreements between human-based and model-based evaluations, to ensure the reliability of model-based evaluations. As shown in Table~\ref{tab:agreement}, we further confirm that agreements between humans and models are high, indicating that model-based evaluations are a reasonable proxy to judge research idea generation. 

\paragraph{Analysis of Refinement Steps}
To see the effectiveness of iterative refinements of research ideas with ReviewingAgents, in Figure~\ref{fig:iteration}, we report the averaged scores on the generated ideas as a function of refinement steps. We first observe initial improvements in the quality of generated ideas with increased refinement steps. Yet, the performance becomes saturated after three iterations, which may indicate diminishing returns for subsequent iteration steps, which aligns with the pattern observed in agent-based refinement work~\cite{multiagent}.

\paragraph{Ablation on Knowledge Sources}
Recall that the full ResearchAgent is augmented with two different knowledge sources, namely relevant references and entities. To see their individual contribution, we perform an ablation study by either excluding one of the knowledge sources or replacing it with random elements. As shown in Table~\ref{tab:ablation}, each knowledge source contributes to performance improvement, and the relevant references are especially helpful. We also note that providing random elements is more helpful than providing no elements at all; we hypothesize that this may be due to the LLM's capability to filter out noise while still gaining incidental value from random inputs.

\begin{figure}[t!]
    \centering
    \includegraphics[width=0.975\columnwidth]{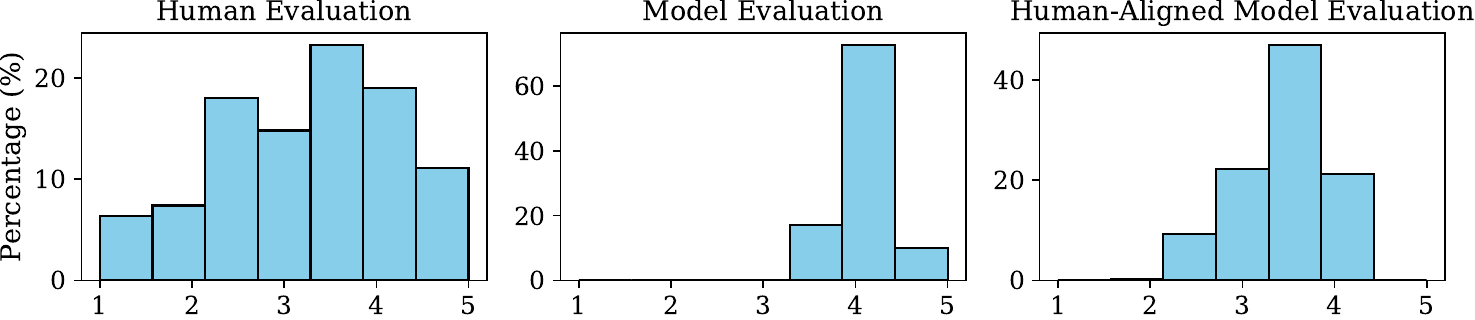}
    \vspace{-0.1in}
    \caption{Distributions of model-based evaluation results with and without the human-induced score criteria alignment (middle and right), as well as human evaluation results (left).}
    \label{fig:distribution}
    \vspace{-0.05in}
\end{figure}
\begin{figure}[t!]
    \centering
    \includegraphics[width=0.975\columnwidth]{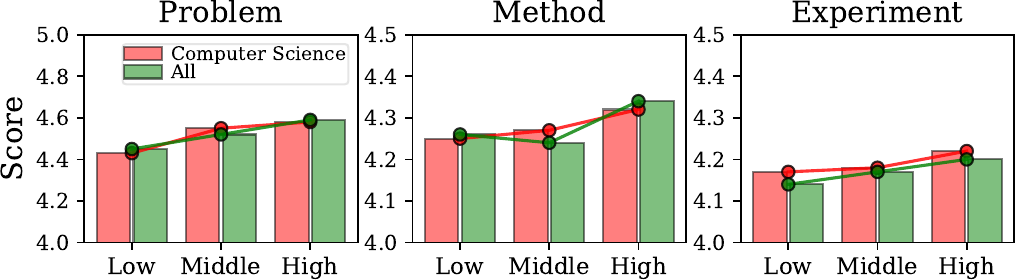}
    \vspace{-0.1in}
    \caption{Results with bucketing papers based on citations.}
    \label{fig:citation}
    \vspace{-0.075in}
\end{figure}

\paragraph{Analysis on Human Alignment for Evaluation}
Recall that to align judgments from model-based evaluations with actual human preferences, we generated the evaluation criteria based on human evaluation results and used them as the criteria for model-based evaluations. Figure~\ref{fig:distribution} demonstrates the efficacy of this strategy, presenting the score distribution of human evaluation compared with the distributions of model-based evaluations with and without human alignment. We find that the score distribution of model-based evaluations without alignment is skewed and different from the score distribution of human judgments. Meanwhile, after aligning the model-based evaluations with human-induced score criteria, the calibrated distribution more closely resembles the distribution of humans.

\paragraph{Correlation on Citation Counts}
We further investigate whether a high-impact paper (when used as a core paper) leads to high-quality research ideas. To measure this, we categorize papers by their citation count (as a proxy for impact), and visualize the average score of each bucket (with model-based evaluations) in Figure~\ref{fig:citation}. We find that ideas from high-impact papers tend to be of higher quality, likely due to their ability to identify research gaps, propose feasible methods, and connect with other works. Additionally, based on the paper distribution (See Figure~\ref{fig:discipline}) and for the ease of manual quality check, evaluation criteria for model-based evaluations are induced mainly with computer science papers. To see whether those criteria are applicable to diverse fields, we also compare a correlation between scores of computer science papers and all papers in Figure~\ref{fig:citation}. From this, we observe that the scores increase when the citation increases for both domains, which may support the generalizability of human-preference-induced evaluation criteria.

\begin{table}[t!]
\caption{Comparisons of ResearchAgent with hypothesis generation methods~\cite{CLBD, tomato}.}
\vspace{-0.1in}
\label{tab:hypothesis}
\small
\centering
\resizebox{0.475\textwidth}{!}{
\renewcommand{\arraystretch}{0.8}
\renewcommand{\tabcolsep}{0.75mm}
\begin{tabular}{lccccc}
\toprule

\textbf{Methods} & \textbf{Clarity} & \textbf{Relevance} & \textbf{Originality} & \textbf{Feasibility} & \textbf{Significance} \\

\midrule
\midrule

SciMON & 4.04 & 4.37 & 4.56 & 3.98 & 4.15 \\

Hypothesis Proposer & 3.97 & 4.14 & 4.07 & 4.01 & 4.11 \\

\noalign{\vskip 0.25ex}\cdashline{1-6}\noalign{\vskip 0.75ex}

ResearchAgent & \textbf{4.11} & \textbf{4.88} & \textbf{4.77} & \textbf{4.05} & \textbf{4.81} \\

\bottomrule

\end{tabular}
}
\vspace{-0.1in}
\end{table}

\paragraph{Comparisons to Hypothesis Generation}
Recall that existing methods for hypothesis generation focus on predicting links between variables or generating hypotheses based on these links, which differs from our experimental setup of generating open-ended research ideas (problems, methods, and experiments). Nevertheless, to understand how the quality of the generated research ideas from prior works~\cite{CLBD, tomato} differs from our ResearchAgent, we perform comparisons. As shown in Table~\ref{tab:hypothesis}, we observe that ResearchAgent is capable of generating superior research hypotheses, due to the utilization of broad and deep knowledge across domains as well as the iterative review and refinement procedures.

\paragraph{Analysis using Different LLMs}
To assess how ResearchAgent's performance changes with different LLMs, we conduct an auxiliary analysis with Llama3, Mixtral, Qwen1.5, and GPT-3.5~\cite{Qwen, Mixtral}, as shown in Table~\ref{tab:gpt35}. These results show a significant performance drop with less capable models. Moreover, the performance differences between the Naive ResearchAgent without knowledge augmentation and the full ResearchAgent become marginal (for Mixtral and GPT-3.5), which indicates that they might struggle with capturing complex concepts between scientific papers. This can likely be attributed to the emergent abilities of LLMs for complex reasoning (but not in smaller LMs)~\cite{Wei2022EmergentAO}, although other subtle issues may also be contributing factors.

\paragraph{Qualitative Analysis}
We provide qualitative results on generated research ideas in Table~\ref{tab:examples}. One representative example in the last row highlights the advantage of entity-centric knowledge augmentation, where two entities (such as Drosophila Genetic Reference Panel and CRISPR) retrieved from the entity-centric knowledge store enable the generation of a novel research idea: bridging genetic variability and CRISPR applications. This exemplifies how external entity-based knowledge uncovers non-trivial relations between scientific concepts.

\begin{table}[t!]
\caption{Results with different, open and proprietary LLMs.}
\vspace{-0.1in}
\label{tab:gpt35}
\small
\centering
\resizebox{0.475\textwidth}{!}{
\renewcommand{\arraystretch}{1.25}
\begin{tabular}{llccc}
\toprule

\textbf{LLMs} & \textbf{Models} & \textbf{Problem} & \textbf{Method} & \textbf{Experiment} \\

\midrule
\midrule

\multirowcell{2}[-0.0ex][l]{\textbf{GPT-4.0}} 

& Naive ResearchAgent & 4.20 & 4.03 & 3.92 \\

& ResearchAgent (Ours) & 4.52 & 4.28 & 4.18 \\

\midrule

\multirowcell{2}[-0.0ex][l]{\textbf{GPT-3.5}} 

& Naive ResearchAgent & 3.56 & 3.56 & 3.63 \\

& ResearchAgent (Ours) & 3.58 & 3.58 & 3.60 \\

\midrule

\multirowcell{2}[-0.0ex][l]{\textbf{Llama3 (8B)}} 

& Naive ResearchAgent & 3.76 & 3.69 & 3.54 \\

& ResearchAgent (Ours) & 4.18 & 4.03 & 3.95 \\

\midrule

\multirowcell{2}[-0.0ex][l]{\textbf{Mixtral (8x7B)}} 

& Naive ResearchAgent & 3.31 & 3.27 & 3.20 \\

& ResearchAgent (Ours) & 3.28 & 3.35 & 3.31 \\

\midrule

\multirowcell{2}[-0.0ex][l]{\textbf{Qwen1.5 (32B)}} 

& Naive ResearchAgent & 3.64 & 3.74 & 3.66 \\

& ResearchAgent (Ours) & 4.02 & 3.97 & 3.94 \\

\bottomrule

\end{tabular}
}
\vspace{-0.1in}
\end{table}

\section{Conclusion}
In this work, we introduced ResearchAgent, a system designed to assist researchers by generating research ideas, which encompass problem identification, method development, and experiment design. Inspired by the human process of ideation, our approach conducts broad and deep literature reviews, integrates knowledge across domains to foster idea cross-pollination, and employs a community of reviewing agents to iteratively refine the generated ideas. Our evaluations, both human and model-based, demonstrated that ResearchAgent produces ideas that are more creative, valid, and clear compared to baselines. While this initial foray shows promising results, multiple challenges remain to operationalize ResearchAgent in real-world research settings. Practical considerations include scaling the knowledge store to encompass diverse research domains, and keeping it current with the latest publications, through which the system can become adaptable even to emerging fields.

\section*{Limitations}
ResearchAgent has some limitations that we hope to address in future work.

First, recall that we built the entity-centric knowledge store to propose beneficial entities during idea generation; however this store is constructed by extracting entities from the titles and abstracts of a limited number of publications (due to the costs of processing them) thereby precluding a large number of other entities and their interconnectedness. 

In addition, the number of entities that we obtain from the BLINK entity linker~\cite{blink} amounts to 3 per paper on average, indicating limited coverage (it is an open-domain linker after all), although it does exhibit generally strong understanding of scientific contexts, as demonstrated by the improvement achieved by the inclusion of its predictions (See Figures~\ref{fig:main} and~\ref{fig:pairwise}, and Table~\ref{tab:examples}). 

Furthermore, since our ResearchAgent is powered by LLMs, similar to any other approaches based on LLMs, it may hallucinate the generated research ideas. While our proposed ResearchAgent can partially mitigate this problem by augmenting LLMs with additional elements, such as references to the target paper and greater entity-centric knowledge, which help ground the generation process in more accurate and relevant information, validating these generated research ideas with experiments is essential to truly accelerate scientific research.

Moreover, while our iterative refinement process with ReviewingAgents demonstrates promising results, it has inherent limitations in scope. Although we employed diverse perspectives by utilizing 15 ReviewingAgents to evaluate three different ideas (problem, method, and experimental design) with five specific criteria for each, this approach may not fully capture the broad range of potential perspectives and criteria necessary for comprehensive evaluation across all different research domains. As acknowledged in the paper, our selection of criteria was informed by their presumed importance, but conducting an exhaustive exploration of all possible criteria over diverse domains is beyond the scope of this work (given the complexity of categorizing and balancing all relevant factors and perspectives). However, we believe the potential of our modular approach allows for customizing and aligning updated or even new criteria to any specific target domain with novel applications, and we leave further expanding them as future work.

Lastly, our ResearchAgent may be less suited for generating ideas in certain domains, such as theoretical sciences, where mathematical reasoning and proof generation play a central role. However, its flexibility allows for customization through specific instructions, enabling the integration of reasoning-based models and techniques to cater to the needs of theoretical research. For instance, in theoretical mathematics, we can instruct (reasoning-based) LLMs to focus on generating proofs or methods and omit experimental design steps that are less relevant. This as an exciting area for future work, where specialized techniques tailored to each domain could be included to broaden its applicability.

\section*{Ethics Statement}
We are aware that the ResearchAgent may have the potential to be misused for nefarious purposes, such as generating research ideas about new explosives, malicious software, and invasive surveillance tools. Notably, this vulnerability is not unique to our approach but a common challenge faced by existing LLMs that possess significant creative and reasoning capabilities, occasionally generating content that may be deemed undesirable. Consequently, it underscores the necessity to enhance the robustness and safety of LLMs more broadly. 

Also, we recognize the risks of unintentional plagiarism associated with using ResearchAgent, where the system might generate ideas that closely mirror existing research due to the regurgitation of training data. While mitigation strategies, such as integrating access to a comprehensive knowledge base to inform users of prior work, can be employed, we understand that building and maintaining such a resource is inherently complex and may not fully eliminate the risk. To further reduce the possibility of plagiarism, recording and tracking all generated ideas could help identify similarities and guide the model to avoid repetition, though this approach would necessitate explicit user consent.

\section*{Acknowledgements}
This work was supported by the National Research Foundation of Korea (NRF) grant funded by the Korea government (MSIT) (No. RS-2023-00256259), the grant of the Korea Machine Learning Ledger Orchestration for Drug Discovery Project (K-MELLODDY) funded by the Ministry of Health \& Welfare and Ministry of Science and ICT, Republic of Korea (grant number: RS-2024-12345678), the Artificial intelligence industrial convergence cluster development project funded by the Ministry of Science and ICT (MSIT, Korea) \& Gwangju Metropolitan City, and the Institute for Information \& communications Technology Planning \& Evaluation (IITP) grant funded by the Korea government (MSIT) (RS-2019-II190075, Artificial Intelligence Graduate School Program (KAIST)).

\bibliography{custom}

\clearpage

\appendix

\begin{figure}[t!]
    \centering
    \includegraphics[width=0.85\columnwidth]{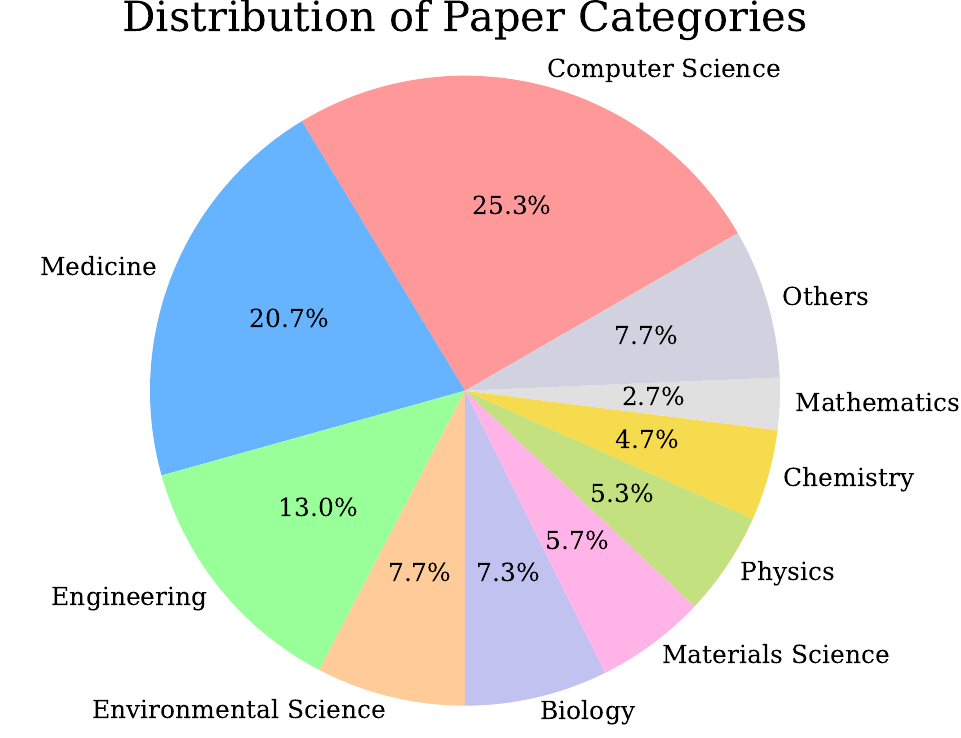}
    \vspace{-0.075in}
    \caption{Visualization of the distribution of disciplines for all core papers, selected for research idea generation.}
    \label{fig:discipline}
    \vspace{-0.1in}
\end{figure}

\section{Additional Experimental Details}
\label{appendix:setups}

In this section, we provide additional details on experiments, including datasets, human evaluation setups, prompts (used for research idea generation and validation), and human-induced criteria.

\subsection{Data Statistics}
We visualize a distribution of core paper categories used for idea generation in Figure~\ref{fig:discipline}, where the categories are obtained from Semantic Scholar API\footnote{https://www.semanticscholar.org/product/api}. From this, we find that the top 3 categories are computer science, medicine, and engineering. 

\subsection{Details on Human Evaluation}
To conduct evaluations with human judges, we recruited 10 researchers from the United States and South Korea, majoring in computer science, medicine, and biology, each with a minimum of 3 published papers. For annotation, they were provided with a 6-page guideline document, which includes the task instruction and annotation examples. After reading this document, the annotators access the Label Studio platform, on which they first read the title and abstract of the target paper, and then review and evaluate the generated research ideas from different models. During the evaluation process, they are allowed to use any external tools, such as web searches. We note that they were compensated at a rate of $\$22.20$ per hour. Also, on average, for one hour, they evaluated 3 sets of research ideas (that are generated from their own papers), with each set comprising three sub-ideas (the problem, method, and experiment design) from three different approaches (i.e., a total of 9 ideas for one hour). We perform three rounds of human evaluations with refinements in between, and, due to the cost associated with human annotations, we are able to fully evaluate a total of 150 ideas.

\begin{figure}[t!]
    \centering
    \includegraphics[width=0.975\columnwidth]{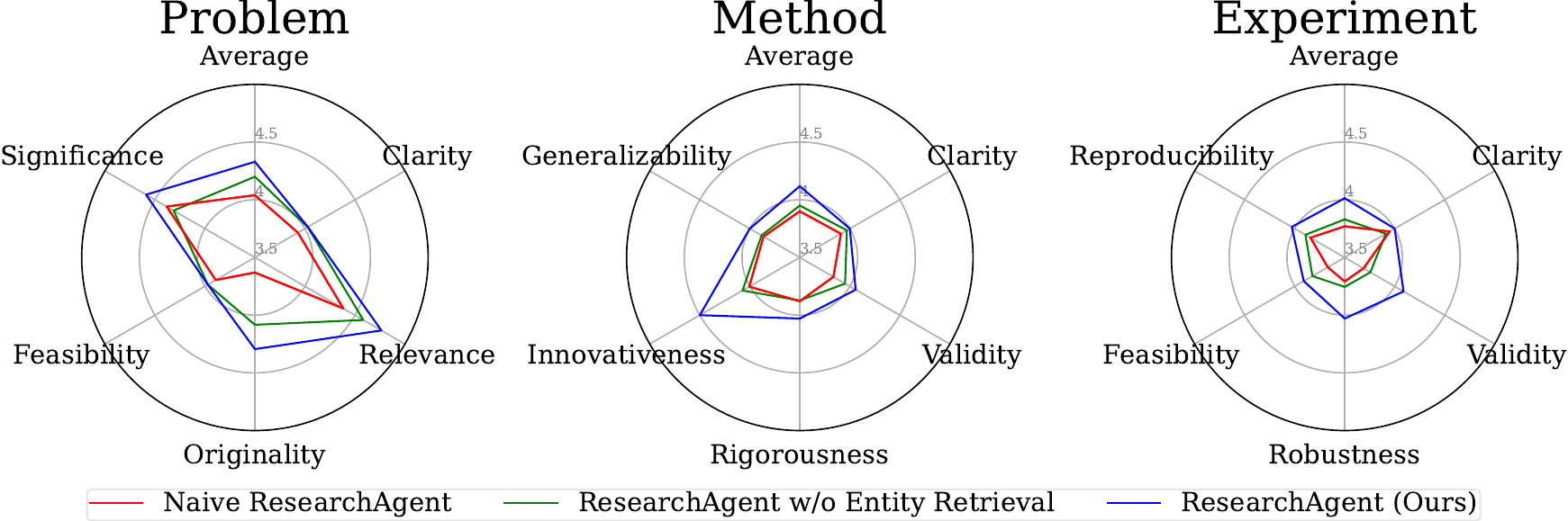}
    \vspace{-0.1in}
    \caption{Results on our research idea generation task with model-based evaluation, where we exclude refinement steps.}
    \label{fig:norefine}
    \vspace{-0.1in}
\end{figure}

\subsection{Prompts for Ideas Generation}
\label{appendix:prompts:generation}

We provide the prompts used to elicit the idea generations from our full ResearchAgent, specifically for instantiating problem identification, method development, and experiment design in Table~\ref{tab:prompt:problem}, Table~\ref{tab:prompt:method}, and Table~\ref{tab:prompt:experiment}, respectively.

\subsection{Prompts for Idea Validation}
\label{appendix:prompts:validation}

We provide the prompts used to elicit the idea validation from our ReviewingAgents as well as the model-based evaluations, specifically for instantiating problem validation, method validation, and experiment design validation in Table~\ref{tab:prompt:problem:valid}, Table~\ref{tab:prompt:method:valid}, and Table~\ref{tab:prompt:experiment:valid}, respectively. In addition, we provide the criteria used, which are induced by human judgments in the next subsection (Appendix~\ref{appendix:criteria}).

\subsection{Criteria Induced by Human Judgements}
\label{appendix:criteria}

Recall that, to align model-based evaluations with human preferences, we induce the criteria (used for automatic evaluations) with actual human judgments. We note that this is done by prompting GPT-4 with 10 pairs of generated ideas and (randomly selected) human judgments. We provide the resulting criteria for validations of problems, methods, and experiment designs in Table~\ref{tab:criteria:problem}, Table~\ref{tab:criteria:method}, and Table~\ref{tab:criteria:experiment}, respectively.

\section{Additional Experimental Results}
We provide additional experimental results, including comparisons without refinements and examples of the generated research ideas.

\subsection{Results without Refinement Steps}
To see whether the proposed ResearchAgent is consistently effective even without ReviewingAgents, we show the model-based evaluation results without any refinement steps in Figure~\ref{fig:norefine}. From this, we clearly observe that the full ResearchAgent outperforms its variants, demonstrating its effectiveness.

\begin{figure*}[t!]
    \centering
    \includegraphics[width=0.9\linewidth]{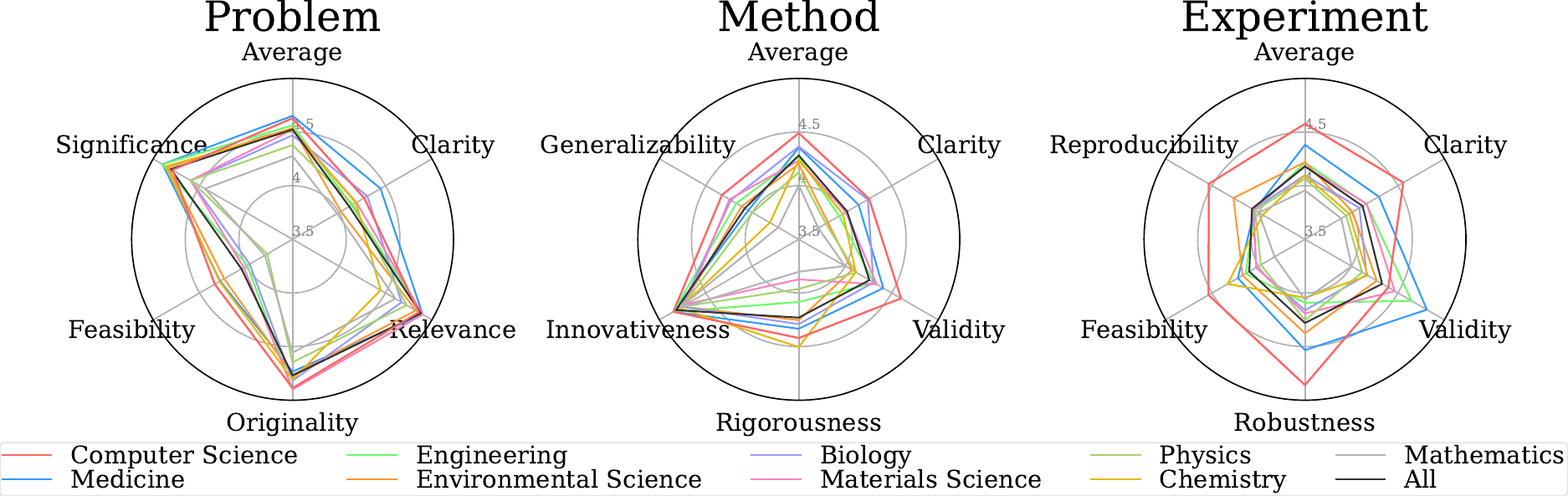}
    \vspace{-0.1in}
    \caption{Breakdown results of the research ideas generated from our full ResearchAgent across different domains.}
    \label{fig:domain}
    \vspace{-0.075in}
\end{figure*}

\subsection{Results on Generated Ideas by Domain}
To see the quality of the generated research ideas across different domains, we breakdown the performance of ResearchAgent according to the categories of core papers in Figure~\ref{fig:discipline}, and present the results in Figure~\ref{fig:domain}. From this, we observe that the generated research ideas on the high-resource domains (such as Computer Science, Medicine, and Engineering where there is a greater volume of existing literature as shown in Figure~\ref{fig:discipline}) are superior to those generated from the low-resource domain papers (such as Physics, Chemistry, and Mathematics). This disparity might be attributed to the fact that the underlying LLMs used to generate research ideas are likely trained on data predominantly sourced from high-resource domains, which leads to enhancing their ability to comprehend scientific concepts and produce relevant research ideas.

\begin{table}[t!]
\caption{Results with different entity retrieval strategies.}
\vspace{-0.1in}
\label{tab:retrieval}
\small
\centering
\resizebox{0.475\textwidth}{!}{
\renewcommand{\arraystretch}{1.15}
\begin{tabular}{lccc}
\toprule

\textbf{Methods} & \textbf{Problem} & \textbf{Method} & \textbf{Experiment} \\

\midrule
\midrule

ResearchAgent &  \\

\noalign{\vskip 0.25ex}\cdashline{1-4}\noalign{\vskip 0.75ex}

- w/ Co-occurrence-based Retrieval & \textbf{4.52 }& 4.28 & \textbf{4.18} \\

- w/ Embedding-based Retrieval & 4.49 & \textbf{4.34} & 4.16 \\

- w/o Entity Retrieval & 4.35 & 4.13 & 4.02 \\

\bottomrule

\end{tabular}
}
\vspace{-0.05in}
\end{table}

\subsection{Analysis with Different Entity Retrieval}
\label{appendix:retrieval}

To see the effectiveness of different entity retrieval strategies, we perform additional experiments, replacing the co-occurrence-based entity retrieval in Equation~\ref{eq:retrieval:detail} to the contextual embedding-based retrieval. Notably, this contextual embedding-based retrieval approach uses the entities that have the highest similarity to the entities appearing in the current literature (i.e., core paper and its references) used for idea generation, where the similarities are calculated based on embedding-level similarities between entities over the latent space represented by the entity linker~\cite{blink}. Therefore, unlike the previous co-occurrence-based entity retrieval that may retrieve entities that have opposite concepts to the main idea of the current core paper (since we often mention limitations of previous work along with the proposed ideas), this embedding-based approach may provide the ResearchAgent with mostly the entities having similar concepts to the core paper. Nevertheless, as shown in Table~\ref{tab:retrieval}, the results with the strategy of entity co-occurrence-based retrieval are comparable to the results with the new embedding-based contextual retrieval. These results might confirm that there is not much difference in the quality of entity retrieval among those two strategies, i.e., most entities retrieved from the co-occurrence-based retrieval are contextually relevant for generating research ideas.

\begin{table*}
    \caption{The prompt used in the full instantiation of ResearchAgent for problem identification.}
    \label{tab:prompt:problem}
    \vspace{-0.1in}
    \resizebox{0.95\textwidth}{!}{
    \renewcommand{\arraystretch}{1.1}
    \renewcommand{\tabcolsep}{2.5mm}
        \begin{tabular}{ll}
        \toprule
        \multicolumn{1}{p{.18\textwidth}}{\textbf{Types}} & \makecell{\multicolumn{1}{p{.92\textwidth}}{\textbf{Texts}}} \\
        \midrule
        \multicolumn{1}{p{.18\textwidth}}{\textbf{System Message}} & 
        \makecell{
            \multicolumn{1}{p{.92\textwidth}}{You are an AI assistant whose primary goal is to identify promising, new, and key scientific problems based on existing scientific literature, in order to aid researchers in discovering novel and significant research opportunities that can advance the field.}
        } \\
        \midrule
        \multicolumn{1}{p{.18\textwidth}}{\textbf{User Message}} & 
        \makecell{
            \multicolumn{1}{p{.92\textwidth}}{{You are going to generate a research problem that should be original, clear, feasible, relevant, and significant to its field. This will be based on the title and abstract of the target paper, those of \{len(references)\} related papers in the existing literature, and \{len(entities)\} entities potentially connected to the research area.}} \\ \\
            \multicolumn{1}{p{.92\textwidth}}{{Understanding of the target paper, related papers, and entities is essential:}} \\
            \multicolumn{1}{p{.92\textwidth}}{{- The target paper is the primary research study you aim to enhance or build upon through future research, serving as the central source and focus for identifying and developing the specific research problem.}} \\
            \multicolumn{1}{p{.92\textwidth}}{{- The related papers are studies that have cited the target paper, indicating their direct relevance and connection to the primary research topic you are focusing on, and providing additional context and insights that are essential for understanding and expanding upon the target paper.}} \\
            \multicolumn{1}{p{.92\textwidth}}{{- The entities can include topics, keywords, individuals, events, or any subjects with possible direct or indirect connections to the target paper or the related studies, serving as auxiliary sources of inspiration or information that may be instrumental in formulating the research problem.}} \\ \\
            \multicolumn{1}{p{.92\textwidth}}{{Your approach should be systematic:}} \\ 
            \multicolumn{1}{p{.92\textwidth}}{{- Start by thoroughly reading the title and abstract of the target paper to understand its core focus.}} \\
            \multicolumn{1}{p{.92\textwidth}}{{- Next, proceed to read the titles and abstracts of the related papers to gain a broader perspective and insights relevant to the primary research topic.}} \\
            \multicolumn{1}{p{.92\textwidth}}{{- Finally, explore the entities to further broaden your perspective, drawing upon a diverse pool of inspiration and information, while keeping in mind that not all may be relevant.}} \\ \\
            \multicolumn{1}{p{.92\textwidth}}{{I am going to provide the target paper, related papers, and entities, as follows: }} \\
            \multicolumn{1}{p{.92\textwidth}}{{Target paper title: \{paper['title']\}}} \\
            \multicolumn{1}{p{.92\textwidth}}{{Target paper abstract: \{paper['abstract']\} }} \\
            \multicolumn{1}{p{.92\textwidth}}{{Related paper titles: \{relatedPaper['titles']\}}} \\
            \multicolumn{1}{p{.92\textwidth}}{{Related paper abstracts: \{relatedPaper['abstracts']\} }} \\
            \multicolumn{1}{p{.92\textwidth}}{{Entities: \{Entities\} }} \\ \\
            \multicolumn{1}{p{.92\textwidth}}{{With the provided target paper, related papers, and entities, your objective now is to formulate a research problem that not only builds upon these existing studies but also strives to be original, clear, feasible, relevant, and significant. Before crafting the research problem, revisit the title and abstract of the target paper, to ensure it remains the focal point of your research problem identification process.}} \\ \\
            \multicolumn{1}{p{.92\textwidth}}{{Target paper title: \{paper['title']\}}} \\
            \multicolumn{1}{p{.92\textwidth}}{{Target paper abstract: \{paper['abstract']\} }} \\ \\
            \multicolumn{1}{p{.92\textwidth}}{{Then, following your review of the above content, please proceed to generate one research problem with the rationale, in the format of }} \\
            \multicolumn{1}{p{.92\textwidth}}{{Problem: }} \\
            \multicolumn{1}{p{.92\textwidth}}{{Rationale: }} \\
        } \\
        \bottomrule
        \end{tabular}
    }
\end{table*}

\begin{table*}
    \caption{The prompt used in the full instantiation of ResearchAgent for method development.}
    \label{tab:prompt:method}
    \vspace{-0.1in}
    \resizebox{0.95\textwidth}{!}{
    \renewcommand{\arraystretch}{1.1}
    \renewcommand{\tabcolsep}{2.5mm}
        \begin{tabular}{ll}
        \toprule
        \multicolumn{1}{p{.18\textwidth}}{\textbf{Types}} & \makecell{\multicolumn{1}{p{.92\textwidth}}{\textbf{Texts}}} \\
        \midrule
        \multicolumn{1}{p{.18\textwidth}}{\textbf{System Message}} & 
        \makecell{
            \multicolumn{1}{p{.92\textwidth}}{You are an AI assistant whose primary goal is to propose innovative, rigorous, and valid methodologies to solve newly identified scientific problems derived from existing scientific literature, in order to empower researchers to pioneer groundbreaking solutions that catalyze breakthroughs in their fields.}
        } \\
        \midrule
        \multicolumn{1}{p{.18\textwidth}}{\textbf{User Message}} & 
        \makecell{
            \multicolumn{1}{p{.92\textwidth}}{{You are going to propose a scientific method to address a specific research problem. Your method should be clear, innovative, rigorous, valid, and generalizable. This will be based on a deep understanding of the research problem, its rationale, existing studies, and various entities.}} \\ \\
            \multicolumn{1}{p{.92\textwidth}}{{Understanding of the research problem, existing studies, and entities is essential:}} \\
            \multicolumn{1}{p{.92\textwidth}}{{- The research problem has been formulated based on an in-depth review of existing studies and a potential exploration of relevant entities, which should be the cornerstone of your method development.}} \\
            \multicolumn{1}{p{.92\textwidth}}{{- The existing studies refer to the target paper that has been pivotal in identifying the problem, as well as the related papers that have been additionally referenced in the problem discovery phase, all serving as foundational material for developing the method.}} \\
            \multicolumn{1}{p{.92\textwidth}}{{- The entities can include topics, keywords, individuals, events, or any subjects with possible direct or indirect connections to the existing studies, serving as auxiliary sources of inspiration or information that may be instrumental in method development.}} \\ \\
            \multicolumn{1}{p{.92\textwidth}}{{Your approach should be systematic:}} \\ 
            \multicolumn{1}{p{.92\textwidth}}{{- Start by thoroughly reading the research problem and its rationale, to understand your primary focus.}} \\
            \multicolumn{1}{p{.92\textwidth}}{{- Next, proceed to review the titles and abstracts of existing studies, to gain a broader perspective and insights relevant to the primary research topic.}} \\
            \multicolumn{1}{p{.92\textwidth}}{{- Finally, explore the entities to further broaden your perspective, drawing upon a diverse pool of inspiration and information, while keeping in mind that not all may be relevant.}} \\ \\
            \multicolumn{1}{p{.92\textwidth}}{{I am going to provide the research problem, existing studies (target paper \& related papers), and entities, as follows: }} \\
            \multicolumn{1}{p{.92\textwidth}}{{Research problem: \{researchProblem\}}} \\
            \multicolumn{1}{p{.92\textwidth}}{{Rationale: \{researchProblemRationale\}}} \\
            \multicolumn{1}{p{.92\textwidth}}{{Target paper title: \{paper['title']\}}} \\
            \multicolumn{1}{p{.92\textwidth}}{{Target paper abstract: \{paper['abstract']\} }} \\
            \multicolumn{1}{p{.92\textwidth}}{{Related paper titles: \{relatedPaper['titles']\}}} \\
            \multicolumn{1}{p{.92\textwidth}}{{Related paper abstracts: \{relatedPaper['abstracts']\} }} \\
            \multicolumn{1}{p{.92\textwidth}}{{Entities: \{Entities\} }} \\ \\
            \multicolumn{1}{p{.92\textwidth}}{{With the provided research problem, existing studies, and entities, your objective now is to formulate a method that not only leverages these resources but also strives to be clear, innovative, rigorous, valid, and generalizable. Before crafting the method, revisit the research problem, to ensure it remains the focal point of your method development process.}} \\ \\
            \multicolumn{1}{p{.92\textwidth}}{{Research problem: \{researchProblem\}}} \\
            \multicolumn{1}{p{.92\textwidth}}{{Rationale: \{researchProblemRationale\}}} \\ \\
            \multicolumn{1}{p{.92\textwidth}}{{Then, following your review of the above content, please proceed to propose your method with its rationale, in the format of }} \\
            \multicolumn{1}{p{.92\textwidth}}{{Method: }} \\
            \multicolumn{1}{p{.92\textwidth}}{{Rationale: }} \\
        } \\
        \bottomrule
        \end{tabular}
    }
\end{table*}

\begin{table*}
    \caption{The prompt used in the full instantiation of ResearchAgent for experiment design.}
    \label{tab:prompt:experiment}
    \vspace{-0.1in}
    \resizebox{0.95\textwidth}{!}{
    \renewcommand{\arraystretch}{1.1}
    \renewcommand{\tabcolsep}{2.5mm}
        \begin{tabular}{ll}
        \toprule
        \multicolumn{1}{p{.18\textwidth}}{\textbf{Types}} & \makecell{\multicolumn{1}{p{.92\textwidth}}{\textbf{Texts}}} \\
        \midrule
        \multicolumn{1}{p{.18\textwidth}}{\textbf{System Message}} & 
        \makecell{
            \multicolumn{1}{p{.92\textwidth}}{You are an AI assistant whose primary goal is to design robust, feasible, and impactful experiments based on identified scientific problems and proposed methodologies from existing scientific literature, in order to enable researchers to systematically test hypotheses and validate groundbreaking discoveries that can transform their respective fields.}
        } \\
        \midrule
        \multicolumn{1}{p{.18\textwidth}}{\textbf{User Message}} & 
        \makecell{
            \multicolumn{1}{p{.92\textwidth}}{{You are going to design an experiment, aimed at validating a proposed method to address a specific research problem. Your experiment design should be clear, robust, reproducible, valid, and feasible. This will be based on a deep understanding of the research problem, scientific method, existing studies, and various entities.}} \\ \\
            \multicolumn{1}{p{.92\textwidth}}{{Understanding of the research problem, scientific method, existing studies, and entities is essential:}} \\
            \multicolumn{1}{p{.92\textwidth}}{{- The research problem has been formulated based on an in-depth review of existing studies and a potential exploration of relevant entities.}} \\
            \multicolumn{1}{p{.92\textwidth}}{{- The scientific method has been proposed to tackle the research problem, which has been informed by insights gained from existing studies and relevant entities.}} \\
            \multicolumn{1}{p{.92\textwidth}}{{- The existing studies refer to the target paper that has been pivotal in identifying the problem and method, as well as the related papers that have been additionally referenced in the discovery phase of the problem and method, all serving as foundational material for designing the experiment.}} \\
            \multicolumn{1}{p{.92\textwidth}}{{- The entities can include topics, keywords, individuals, events, or any subjects with possible direct or indirect connections to the existing studies, serving as auxiliary sources of inspiration or information that may be instrumental in your experiment design.}} \\ \\
            \multicolumn{1}{p{.92\textwidth}}{{Your approach should be systematic:}} \\ 
            \multicolumn{1}{p{.92\textwidth}}{{- Start by thoroughly reading the research problem and its rationale followed by the proposed method and its rationale, to pinpoint your primary focus.}} \\
            \multicolumn{1}{p{.92\textwidth}}{{- Next, proceed to review the titles and abstracts of existing studies, to gain a broader perspective and insights relevant to the primary research topic.}} \\
            \multicolumn{1}{p{.92\textwidth}}{{- Finally, explore the entities to further broaden your perspective, drawing upon a diverse pool of inspiration and information, while keeping in mind that not all may be relevant.}} \\ \\
            \multicolumn{1}{p{.92\textwidth}}{{I am going to provide the research problem, scientific method, existing studies (target paper \& related papers), and entities, as follows: }} \\
            \multicolumn{1}{p{.92\textwidth}}{{Research problem: \{researchProblem\}}} \\
            \multicolumn{1}{p{.92\textwidth}}{{Rationale: \{researchProblemRationale\}}} \\
            \multicolumn{1}{p{.92\textwidth}}{{Scientific method: \{scientificMethod\}}} \\
            \multicolumn{1}{p{.92\textwidth}}{{Rationale: \{scientificMethodRationale\}}} \\
            \multicolumn{1}{p{.92\textwidth}}{{Target paper title: \{paper['title']\}}} \\
            \multicolumn{1}{p{.92\textwidth}}{{Target paper abstract: \{paper['abstract']\} }} \\
            \multicolumn{1}{p{.92\textwidth}}{{Related paper titles: \{relatedPaper['titles']\}}} \\
            \multicolumn{1}{p{.92\textwidth}}{{Related paper abstracts: \{relatedPaper['abstracts']\} }} \\
            \multicolumn{1}{p{.92\textwidth}}{{Entities: \{Entities\} }} \\ \\
            \multicolumn{1}{p{.92\textwidth}}{{With the provided research problem, scientific method, existing studies, and entities, your objective now is to design an experiment that not only leverages these resources but also strives to be clear, robust, reproducible, valid, and feasible. Before crafting the experiment design, revisit the research problem and proposed method, to ensure they remain at the center of your experiment design process.}} \\ \\
            \multicolumn{1}{p{.92\textwidth}}{{Research problem: \{researchProblem\}}} \\
            \multicolumn{1}{p{.92\textwidth}}{{Rationale: \{researchProblemRationale\}}} \\ 
            \multicolumn{1}{p{.92\textwidth}}{{Scientific method: \{scientificMethod\}}} \\
            \multicolumn{1}{p{.92\textwidth}}{{Rationale: \{scientificMethodRationale\}}} \\ \\
            \multicolumn{1}{p{.92\textwidth}}{{Then, following your review of the above content, please proceed to outline your experiment with its rationale, in the format of }} \\
            \multicolumn{1}{p{.92\textwidth}}{{Experiment: }} \\
            \multicolumn{1}{p{.92\textwidth}}{{Rationale: }} \\
        } \\
        \bottomrule
        \end{tabular}
    }
\end{table*}
\begin{table*}
    \caption{The prompt used in the full instantiation of ReviewingAgent for problem validation.}
    \label{tab:prompt:problem:valid}
    \vspace{-0.1in}
    \resizebox{0.95\textwidth}{!}{
    \renewcommand{\arraystretch}{1.1}
    \renewcommand{\tabcolsep}{2.5mm}
        \begin{tabular}{ll}
        \toprule
        \multicolumn{1}{p{.18\textwidth}}{\textbf{Types}} & \makecell{\multicolumn{1}{p{.92\textwidth}}{\textbf{Texts}}} \\
        \midrule
        \multicolumn{1}{p{.18\textwidth}}{\textbf{System Message}} & 
        \makecell{
            \multicolumn{1}{p{.92\textwidth}}{You are an AI assistant whose primary goal is to assess the quality and validity of scientific problems across diverse dimensions, in order to aid researchers in refining their problems based on your evaluations and feedback, thereby enhancing the impact and reach of their work.}
        } \\
        \midrule
        \multicolumn{1}{p{.18\textwidth}}{\textbf{User Message}} & 
        \makecell{
            \multicolumn{1}{p{.92\textwidth}}{{You are going to evaluate a research problem for its \{metric\}, focusing on how well it is defined in a clear, precise, and understandable manner.}} \\ \\
            \multicolumn{1}{p{.92\textwidth}}{{As part of your evaluation, you can refer to the existing studies that may be related to the problem, which will help in understanding the context of the problem for a more comprehensive assessment.}} \\
            \multicolumn{1}{p{.92\textwidth}}{{- The existing studies refer to the target paper that has been pivotal in identifying the problem, as well as the related papers that have been additionally referenced in the discovery phase of the problem.}} \\ \\
            \multicolumn{1}{p{.92\textwidth}}{{The existing studies (target paper \& related papers) are as follows:}} \\ 
            \multicolumn{1}{p{.92\textwidth}}{{Target paper title: \{paper['title']\}}} \\
            \multicolumn{1}{p{.92\textwidth}}{{Target paper abstract: \{paper['abstract']\} }} \\
            \multicolumn{1}{p{.92\textwidth}}{{Related paper titles: \{relatedPaper['titles']\}}} \\
            \multicolumn{1}{p{.92\textwidth}}{{Related paper abstracts: \{relatedPaper['abstracts']\} }} \\ \\
            \multicolumn{1}{p{.92\textwidth}}{{Now, proceed with your \{metric\} evaluation approach that should be systematic:}} \\
            \multicolumn{1}{p{.92\textwidth}}{{- Start by thoroughly reading the research problem and its rationale, keeping in mind the context provided by the existing studies mentioned above.}} \\
            \multicolumn{1}{p{.92\textwidth}}{{- Next, generate a review and feedback that should be constructive, helpful, and concise, focusing on the \{metric\} of the problem.}} \\
            \multicolumn{1}{p{.92\textwidth}}{{- Finally, provide a score on a 5-point Likert scale, with 1 being the lowest, please ensuring a discerning and critical evaluation to avoid a tendency towards uniformly high ratings (4-5) unless fully justified:}} \\ 
            \multicolumn{1}{p{.92\textwidth}}{{\{criteria\}}} \\ \\
            \multicolumn{1}{p{.92\textwidth}}{{I am going to provide the research problem with its rationale, as follows:}} \\
            \multicolumn{1}{p{.92\textwidth}}{{Research problem: \{researchProblem\}}} \\
            \multicolumn{1}{p{.92\textwidth}}{{Rationale: \{researchProblemRationale\}}} \\ \\
            \multicolumn{1}{p{.92\textwidth}}{{After your evaluation of the above content, please provide your review, feedback, and rating, in the format of}} \\ 
            \multicolumn{1}{p{.92\textwidth}}{{Review: }} \\
            \multicolumn{1}{p{.92\textwidth}}{{Feedback: }} \\
            \multicolumn{1}{p{.92\textwidth}}{{Rating (1-5): }} \\
        } \\
        \bottomrule
        \end{tabular}
    }
\end{table*}

\begin{table*}
    \caption{The prompt used in the full instantiation of ReviewingAgent for method validation.}
    \label{tab:prompt:method:valid}
    \vspace{-0.1in}
    \resizebox{0.95\textwidth}{!}{
    \renewcommand{\arraystretch}{1.1}
    \renewcommand{\tabcolsep}{2.5mm}
        \begin{tabular}{ll}
        \toprule
        \multicolumn{1}{p{.18\textwidth}}{\textbf{Types}} & \makecell{\multicolumn{1}{p{.92\textwidth}}{\textbf{Texts}}} \\
        \midrule
        \multicolumn{1}{p{.18\textwidth}}{\textbf{System Message}} & 
        \makecell{
            \multicolumn{1}{p{.92\textwidth}}{You are an AI assistant whose primary goal is to assess the quality and soundness of scientific methods across diverse dimensions, in order to aid researchers in refining their methods based on your evaluations and feedback, thereby enhancing the impact and reach of their work.}
        } \\
        \midrule
        \multicolumn{1}{p{.18\textwidth}}{\textbf{User Message}} & 
        \makecell{
            \multicolumn{1}{p{.92\textwidth}}{{You are going to evaluate a scientific method for its \{metric\} in addressing a research problem, focusing on how well it is described in a clear, precise, and understandable manner that allows for replication and comprehension of the approach.}} \\ \\
            \multicolumn{1}{p{.92\textwidth}}{{As part of your evaluation, you can refer to the research problem, and existing studies, which will help in understanding the context of the proposed method for a more comprehensive assessment.}} \\
            \multicolumn{1}{p{.92\textwidth}}{{- The research problem has been used as the cornerstone of the method development, formulated based on an in-depth review of existing studies and a potential exploration of relevant entities.}} \\
            \multicolumn{1}{p{.92\textwidth}}{{- The existing studies refer to the target paper that has been pivotal in identifying the problem and method, as well as the related papers that have been additionally referenced in the discovery phase of the problem and method.}} \\ \\
            \multicolumn{1}{p{.92\textwidth}}{{The research problem and existing studies (target paper \& related papers) are as follows:}} \\ 
            \multicolumn{1}{p{.92\textwidth}}{{Research problem: \{researchProblem\}}} \\
            \multicolumn{1}{p{.92\textwidth}}{{Rationale: \{researchProblemRationale\}}} \\ 
            \multicolumn{1}{p{.92\textwidth}}{{Target paper title: \{paper['title']\}}} \\
            \multicolumn{1}{p{.92\textwidth}}{{Target paper abstract: \{paper['abstract']\} }} \\
            \multicolumn{1}{p{.92\textwidth}}{{Related paper titles: \{relatedPaper['titles']\}}} \\
            \multicolumn{1}{p{.92\textwidth}}{{Related paper abstracts: \{relatedPaper['abstracts']\} }} \\ \\
            \multicolumn{1}{p{.92\textwidth}}{{Now, proceed with your \{metric\} evaluation approach that should be systematic:}} \\
            \multicolumn{1}{p{.92\textwidth}}{{- Start by thoroughly reading the proposed method and its rationale, keeping in mind the context provided by the research problem, and existing studies mentioned above.}} \\
            \multicolumn{1}{p{.92\textwidth}}{{- Next, generate a review and feedback that should be constructive, helpful, and concise, focusing on the \{metric\} of the method.}} \\
            \multicolumn{1}{p{.92\textwidth}}{{- Finally, provide a score on a 5-point Likert scale, with 1 being the lowest, please ensuring a discerning and critical evaluation to avoid a tendency towards uniformly high ratings (4-5) unless fully justified:}} \\ 
            \multicolumn{1}{p{.92\textwidth}}{{\{criteria\}}} \\ \\
            \multicolumn{1}{p{.92\textwidth}}{{I am going to provide the proposed method with its rationale, as follows:}} \\
            \multicolumn{1}{p{.92\textwidth}}{{Scientific method: \{scientificMethod\}}} \\
            \multicolumn{1}{p{.92\textwidth}}{{Rationale: \{scientificMethodRationale\}}} \\ \\
            \multicolumn{1}{p{.92\textwidth}}{{After your evaluation of the above content, please provide your review, feedback, and rating, in the format of}} \\ 
            \multicolumn{1}{p{.92\textwidth}}{{Review: }} \\
            \multicolumn{1}{p{.92\textwidth}}{{Feedback: }} \\
            \multicolumn{1}{p{.92\textwidth}}{{Rating (1-5): }} \\
        } \\
        \bottomrule
        \end{tabular}
    }
\end{table*}

\begin{table*}
    \caption{The prompt used in the full instantiation of ReviewingAgent for experiment design validation.}
    \label{tab:prompt:experiment:valid}
    \vspace{-0.1in}
    \resizebox{0.95\textwidth}{!}{
    \renewcommand{\arraystretch}{1.1}
    \renewcommand{\tabcolsep}{2.5mm}
        \begin{tabular}{ll}
        \toprule
        \multicolumn{1}{p{.18\textwidth}}{\textbf{Types}} & \makecell{\multicolumn{1}{p{.92\textwidth}}{\textbf{Texts}}} \\
        \midrule
        \multicolumn{1}{p{.18\textwidth}}{\textbf{System Message}} & 
        \makecell{
            \multicolumn{1}{p{.92\textwidth}}{You are an AI assistant whose primary goal is to meticulously evaluate the experimental designs of scientific papers across diverse dimensions, in order to aid researchers in refining their experimental approaches based on your evaluations and feedback, thereby amplifying the quality and impact of their scientific contributions.}
        } \\
        \midrule
        \multicolumn{1}{p{.18\textwidth}}{\textbf{User Message}} & 
        \makecell{
            \multicolumn{1}{p{.92\textwidth}}{{You are going to evaluate an experiment design for its \{metric\} in validating a scientific method to address a research problem, focusing on how well it is described in a clear, precise, and understandable manner, enabling others to grasp the setup, procedure, and expected outcomes.}} \\ \\
            \multicolumn{1}{p{.92\textwidth}}{{As part of your evaluation, you can refer to the research problem, scientific method, and existing studies, which will help in understanding the context of the designed experiment for a more comprehensive assessment.}} \\
            \multicolumn{1}{p{.92\textwidth}}{{- The research problem has been formulated based on an in-depth review of existing studies and a potential exploration of relevant entities.}} \\
            \multicolumn{1}{p{.92\textwidth}}{{- The scientific method has been proposed to tackle the research problem, which has been informed by insights gained from existing studies and relevant entities.}} \\ 
            \multicolumn{1}{p{.92\textwidth}}{{- The existing studies refer to the target paper that has been pivotal in identifying the problem, method, and experiment, as well as the related papers that have been additionally referenced in their discovery phases.}} \\ \\
            \multicolumn{1}{p{.92\textwidth}}{{The research problem, scientific method, and existing studies (target paper \& related papers) are as follows:}} \\ 
            \multicolumn{1}{p{.92\textwidth}}{{Research problem: \{researchProblem\}}} \\
            \multicolumn{1}{p{.92\textwidth}}{{Rationale: \{researchProblemRationale\}}} \\ 
            \multicolumn{1}{p{.92\textwidth}}{{Scientific method: \{scientificMethod\}}} \\
            \multicolumn{1}{p{.92\textwidth}}{{Rationale: \{scientificMethodRationale\}}} \\
            \multicolumn{1}{p{.92\textwidth}}{{Target paper title: \{paper['title']\}}} \\
            \multicolumn{1}{p{.92\textwidth}}{{Target paper abstract: \{paper['abstract']\} }} \\
            \multicolumn{1}{p{.92\textwidth}}{{Related paper titles: \{relatedPaper['titles']\}}} \\
            \multicolumn{1}{p{.92\textwidth}}{{Related paper abstracts: \{relatedPaper['abstracts']\} }} \\ \\
            \multicolumn{1}{p{.92\textwidth}}{{Now, proceed with your \{metric\} evaluation approach that should be systematic:}} \\
            \multicolumn{1}{p{.92\textwidth}}{{- Start by thoroughly reading the experiment design and its rationale, keeping in mind the context provided by the research problem, scientific method, and existing studies mentioned above.}} \\
            \multicolumn{1}{p{.92\textwidth}}{{- Next, generate a review and feedback that should be constructive, helpful, and concise, focusing on the \{metric\} of the experiment.}} \\
            \multicolumn{1}{p{.92\textwidth}}{{- Finally, provide a score on a 5-point Likert scale, with 1 being the lowest, please ensuring a discerning and critical evaluation to avoid a tendency towards uniformly high ratings (4-5) unless fully justified:}} \\ 
            \multicolumn{1}{p{.92\textwidth}}{{\{criteria\}}} \\ \\
            \multicolumn{1}{p{.92\textwidth}}{{I am going to provide the designed experiment with its rationale, as follows:}} \\
            \multicolumn{1}{p{.92\textwidth}}{{Experiment design: \{experimentDesign\}}} \\
            \multicolumn{1}{p{.92\textwidth}}{{Rationale: \{experimentDesignRationale\}}} \\ \\
            \multicolumn{1}{p{.92\textwidth}}{{After your evaluation of the above content, please provide your review, feedback, and rating, in the format of}} \\ 
            \multicolumn{1}{p{.92\textwidth}}{{Review: }} \\
            \multicolumn{1}{p{.92\textwidth}}{{Feedback: }} \\
            \multicolumn{1}{p{.92\textwidth}}{{Rating (1-5): }} \\
        } \\
        \bottomrule
        \end{tabular}
    }
\end{table*}

\begin{table*}
    \caption{The criteria used for evaluating research ideas: problems, methods, and experiment designs.}
    \label{tab:criteria}
    \vspace{-0.1in}
    \resizebox{0.95\textwidth}{!}{
    \renewcommand{\arraystretch}{1.1}
    \renewcommand{\tabcolsep}{2.5mm}
        \begin{tabular}{lll}
        \toprule
        \multicolumn{1}{p{.12\textwidth}}{\textbf{Types}} & \makecell{\multicolumn{1}{p{.12\textwidth}}{\textbf{Criteria}}} & \makecell{\multicolumn{1}{p{1.05\textwidth}}{\textbf{Texts}}} \\
        \midrule
        \multirow{8}{*}{\textbf{Problem}} 
        & \makecell{\multicolumn{1}{p{.12\textwidth}}{\textbf{Clarity}}} & \makecell{
            \multicolumn{1}{p{1.05\textwidth}}{It assesses whether the problem is defined in a clear, precise, and understandable manner.}
        } \\
        \noalign{\vskip 0.25ex}\cdashline{2-3}\noalign{\vskip 0.75ex}
        & \makecell{\multicolumn{1}{p{.12\textwidth}}{\textbf{Relevance}}} & \makecell{
            \multicolumn{1}{p{1.05\textwidth}}{It measures whether the problem is pertinent and applicable to the current field or context of study.}
        } \\
        \noalign{\vskip 0.25ex}\cdashline{2-3}\noalign{\vskip 0.75ex}
        & \makecell{\multicolumn{1}{p{.12\textwidth}}{\textbf{Originality}}} & \makecell{
            \multicolumn{1}{p{1.05\textwidth}}{It evaluates whether the problem presents a novel challenge or unique perspective that has not been extensively explored before.}
        } \\
        \noalign{\vskip 0.25ex}\cdashline{2-3}\noalign{\vskip 0.75ex}
        & \makecell{\multicolumn{1}{p{.12\textwidth}}{\textbf{Feasibility}}} & \makecell{
            \multicolumn{1}{p{1.05\textwidth}}{It examines whether the problem can realistically be investigated or solved with the available resources and within reasonable constraints.}
        } \\
        \noalign{\vskip 0.25ex}\cdashline{2-3}\noalign{\vskip 0.75ex}
        & \makecell{\multicolumn{1}{p{.12\textwidth}}{\textbf{Significance}}} & \makecell{
            \multicolumn{1}{p{1.05\textwidth}}{It assesses the importance and potential impact of solving the problem, including its contribution to the field or its broader implications.}
        } \\
        \midrule
        \multirow{10}{*}{\textbf{Method}} 
        & \makecell{\multicolumn{1}{p{.12\textwidth}}{\textbf{Clarity}}} & \makecell{
            \multicolumn{1}{p{1.05\textwidth}}{It assesses whether the method is described in a clear, precise, and understandable manner that allows for replication and comprehension of the approach.}
        } \\
        \noalign{\vskip 0.25ex}\cdashline{2-3}\noalign{\vskip 0.75ex}
        & \makecell{\multicolumn{1}{p{.12\textwidth}}{\textbf{Validity}}} & \makecell{
            \multicolumn{1}{p{1.05\textwidth}}{It measures the accuracy, relevance, and soundness of the method in addressing the research problem, ensuring that it is appropriate and directly relevant to the objectives of the study.}
        } \\
        \noalign{\vskip 0.25ex}\cdashline{2-3}\noalign{\vskip 0.75ex}
        & \makecell{\multicolumn{1}{p{.12\textwidth}}{\textbf{Rigorousness}}} & \makecell{
            \multicolumn{1}{p{1.05\textwidth}}{It examines the thoroughness, precision, and consistency of the method, ensuring that the approach is systematic, well-structured, and adheres to high standards of research quality.}
        } \\
        \noalign{\vskip 0.25ex}\cdashline{2-3}\noalign{\vskip 0.75ex}
        & \makecell{\multicolumn{1}{p{.12\textwidth}}{\textbf{Innovativeness}}} & \makecell{
            \multicolumn{1}{p{1.05\textwidth}}{It evaluates whether the method introduces new techniques, approaches, or perspectives to the research field that differ from standard research practices and advance them in the field.}
        } \\
        \noalign{\vskip 0.25ex}\cdashline{2-3}\noalign{\vskip 0.75ex}
        & \makecell{\multicolumn{1}{p{.12\textwidth}}{\textbf{Generalizability}}} & \makecell{
            \multicolumn{1}{p{1.05\textwidth}}{It assesses the extent to which the method can be applied to or is relevant for other contexts, populations, or settings beyond the scope of the study.}
        } \\
        \midrule
        \multirow{12}{*}{\textbf{Experiment}} 
        & \makecell{\multicolumn{1}{p{.12\textwidth}}{\textbf{Clarity}}} & \makecell{
            \multicolumn{1}{p{1.05\textwidth}}{It determines whether the experiment design is described in a clear, precise, and understandable manner, enabling others to grasp the setup, procedure, and expected outcomes.}
        } \\
        \noalign{\vskip 0.25ex}\cdashline{2-3}\noalign{\vskip 0.75ex}
        & \makecell{\multicolumn{1}{p{.12\textwidth}}{\textbf{Validity}}} & \makecell{
            \multicolumn{1}{p{1.05\textwidth}}{It measures the appropriateness and soundness of the experimental design in accurately addressing the research questions or effectively validating the proposed methods, ensuring that the design effectively tests what it is intended to examine.}
        } \\
        \noalign{\vskip 0.25ex}\cdashline{2-3}\noalign{\vskip 0.75ex}
        & \makecell{\multicolumn{1}{p{.12\textwidth}}{\textbf{Robustness}}} & \makecell{
            \multicolumn{1}{p{1.05\textwidth}}{It evaluates the durability of the experimental design across a wide range of conditions and variables, ensuring that the outcomes are not reliant on a few specific cases and remain consistent across a broad spectrum of scenarios.}
        } \\
        \noalign{\vskip 0.25ex}\cdashline{2-3}\noalign{\vskip 0.75ex}
        & \makecell{\multicolumn{1}{p{.12\textwidth}}{\textbf{Feasibility}}} & \makecell{
            \multicolumn{1}{p{1.05\textwidth}}{It evaluates whether the experiment design can realistically be implemented with the available resources, time, and technological or methodological constraints, ensuring that the experiment is practical and achievable.}
        } \\
        \noalign{\vskip 0.25ex}\cdashline{2-3}\noalign{\vskip 0.75ex}
        & \makecell{\multicolumn{1}{p{.12\textwidth}}{\textbf{Reproducibility}}} & \makecell{
            \multicolumn{1}{p{1.05\textwidth}}{It examines whether the information provided is sufficient and detailed enough for other researchers to reproduce the experiment using the same methodology and conditions, ensuring the reliability of the findings.}
        } \\
        \bottomrule
        \end{tabular}
    }
\end{table*}

\begin{table*}
    \caption{The criteria induced from human judgments for validating the identified problems, which are used to align model-based evaluations with actual human preferences.}
    \label{tab:criteria:problem}
    \vspace{-0.1in}
    \resizebox{0.95\textwidth}{!}{
    \renewcommand{\arraystretch}{1.1}
    \renewcommand{\tabcolsep}{2.5mm}
        \begin{tabular}{lll}
        \toprule
        \multicolumn{1}{p{.12\textwidth}}{\textbf{Types}} & \makecell{\multicolumn{1}{p{.12\textwidth}}{\textbf{Criteria}}} & \makecell{\multicolumn{1}{p{1.05\textwidth}}{\textbf{Texts}}} \\
        \midrule
        \multirow{39}{*}{\textbf{Problem}} 
        & \makecell{\multicolumn{1}{p{.12\textwidth}}{\textbf{Clarity}}} & \makecell{
            \multicolumn{1}{p{1.05\textwidth}}{1. The problem is presented in a highly ambiguous manner, lacking clear definition and leaving significant room for interpretation or confusion.} \\
            \multicolumn{1}{p{1.05\textwidth}}{2. The problem is somewhat defined but suffers from vague terms and insufficient detail, making it challenging to grasp the full scope or objective.} \\
            \multicolumn{1}{p{1.05\textwidth}}{3. The problem is stated in a straightforward manner, but lacks the depth or specificity needed to fully convey the nuances and boundaries of the research scope.} \\
            \multicolumn{1}{p{1.05\textwidth}}{4. The problem is clearly articulated with precise terminology and sufficient detail, providing a solid understanding of the scope and objectives with minimal ambiguity.} \\
            \multicolumn{1}{p{1.05\textwidth}}{5. The problem is exceptionally clear, concise, and specific, with every term and aspect well-defined, leaving no room for misinterpretation and fully encapsulating the research scope and aims.}
        } \\
        \noalign{\vskip 0.25ex}\cdashline{2-3}\noalign{\vskip 0.75ex}
        & \makecell{\multicolumn{1}{p{.12\textwidth}}{\textbf{Relevance}}} & \makecell{
            \multicolumn{1}{p{1.05\textwidth}}{1. The problem shows almost no relevance to the current field, failing to connect with the established context or build upon existing work.} \\
            \multicolumn{1}{p{1.05\textwidth}}{2. The problem has minimal relevance, with only superficial connections to the field and a lack of meaningful integration with prior studies.} \\
            \multicolumn{1}{p{1.05\textwidth}}{3. The problem is somewhat relevant, making a moderate attempt to align with the field but lacking significant innovation or depth.} \\
            \multicolumn{1}{p{1.05\textwidth}}{4. The problem is relevant and well-connected to the field, demonstrating a good understanding of existing work and offering promising contributions.} \\
            \multicolumn{1}{p{1.05\textwidth}}{5. The problem is highly relevant, deeply integrated with the current context, and represents a significant advancement in the field.}
        } \\
        \noalign{\vskip 0.25ex}\cdashline{2-3}\noalign{\vskip 0.75ex}
        & \makecell{\multicolumn{1}{p{.12\textwidth}}{\textbf{Originality}}} & \makecell{
            \multicolumn{1}{p{1.05\textwidth}}{1. The problem exhibits no discernible originality, closely mirroring existing studies without introducing any novel perspectives or challenges.} \\
            \multicolumn{1}{p{1.05\textwidth}}{2. The problem shows minimal originality, with slight variations from known studies, lacking significant new insights or innovative approaches.} \\
            \multicolumn{1}{p{1.05\textwidth}}{3. The problem demonstrates moderate originality, offering some new insights or angles, but these are not sufficiently groundbreaking or distinct from existing work.} \\
            \multicolumn{1}{p{1.05\textwidth}}{4. The problem is notably original, presenting a unique challenge or perspective that is well-differentiated from existing studies, contributing valuable new understanding to the field.} \\
            \multicolumn{1}{p{1.05\textwidth}}{5. The problem is highly original, introducing a pioneering challenge or perspective that has not been explored before, setting a new direction for future research.}
        } \\
        \noalign{\vskip 0.25ex}\cdashline{2-3}\noalign{\vskip 0.75ex}
        & \makecell{\multicolumn{1}{p{.12\textwidth}}{\textbf{Feasibility}}} & \makecell{
            \multicolumn{1}{p{1.05\textwidth}}{1. The problem is fundamentally infeasible due to insurmountable resource constraints, lack of foundational research, or critical methodological flaws.} \\
            \multicolumn{1}{p{1.05\textwidth}}{2. The problem faces significant feasibility challenges related to resource availability, existing knowledge gaps, or technical limitations, making progress unlikely.} \\
            \multicolumn{1}{p{1.05\textwidth}}{3. The problem is feasible to some extent but faces notable obstacles in resources, existing research support, or technical implementation, which could hinder significant advancements.} \\
            \multicolumn{1}{p{1.05\textwidth}}{4. The problem is mostly feasible with manageable challenges in resources, supported by adequate existing research, and has a clear, achievable methodology, though minor issues may persist.} \\
            \multicolumn{1}{p{1.05\textwidth}}{5. The problem is highly feasible with minimal barriers, well-supported by existing research, ample resources, and a robust, clear methodology, promising significant advancements.}
        } \\
        \noalign{\vskip 0.25ex}\cdashline{2-3}\noalign{\vskip 0.75ex}
        & \makecell{\multicolumn{1}{p{.12\textwidth}}{\textbf{Significance}}} & \makecell{
            \multicolumn{1}{p{1.05\textwidth}}{1. The problem shows minimal to no significance, lacking relevance or potential impact in advancing the field or contributing to practical applications.} \\
            \multicolumn{1}{p{1.05\textwidth}}{2. The problem has limited significance, with a narrow scope of impact and minor contributions to the field, offering little to no practical implications.} \\
            \multicolumn{1}{p{1.05\textwidth}}{3. The problem demonstrates average significance, with some contributions to the field and potential practical implications, but lacks innovation or broader impact.} \\
            \multicolumn{1}{p{1.05\textwidth}}{4. The problem is significant, offering notable contributions to the field and valuable practical implications, with evidence of potential for broader impact and advancement.} \\
            \multicolumn{1}{p{1.05\textwidth}}{5. The problem presents exceptional significance, with groundbreaking contributions to the field, broad and transformative potential impacts, and substantial practical applications across diverse domains.}
        } \\
        \bottomrule
        \end{tabular}
    }
\end{table*}

\begin{table*}
    \caption{The criteria induced from human judgments for validating the developed methods, which used to align model-based evaluations with actual human preferences.}
    \label{tab:criteria:method}
    \vspace{-0.1in}
    \resizebox{0.95\textwidth}{!}{
    \renewcommand{\arraystretch}{1.1}
    \renewcommand{\tabcolsep}{2.5mm}
        \begin{tabular}{lll}
        \toprule
        \multicolumn{1}{p{.12\textwidth}}{\textbf{Types}} & \makecell{\multicolumn{1}{p{.12\textwidth}}{\textbf{Criteria}}} & \makecell{\multicolumn{1}{p{1.05\textwidth}}{\textbf{Texts}}} \\
        \midrule
        \multirow{39}{*}{\textbf{Method}} 
        & \makecell{\multicolumn{1}{p{.12\textwidth}}{\textbf{Clarity}}} & \makecell{
            \multicolumn{1}{p{1.05\textwidth}}{1. The method is explained in an extremely vague or ambiguous manner, making it impossible to understand or replicate the approach without additional information or clarification.} \\
            \multicolumn{1}{p{1.05\textwidth}}{2. The method is described with some detail, but significant gaps in explanation or logic leave the reader with considerable confusion and uncertainty about how to apply or replicate the approach.} \\
            \multicolumn{1}{p{1.05\textwidth}}{3. The method is described with sufficient detail to understand the basic approach, but lacks the precision or specificity needed to fully replicate or grasp the nuances of the methodology without further guidance.} \\
            \multicolumn{1}{p{1.05\textwidth}}{4. The method is clearly and precisely described, with most details provided to allow for replication and comprehension, though minor areas may benefit from further clarification or elaboration.} \\
            \multicolumn{1}{p{1.05\textwidth}}{5. The method is articulated in an exceptionally clear, precise, and detailed manner, enabling straightforward replication and thorough understanding of the approach with no ambiguities.}
        } \\
        \noalign{\vskip 0.25ex}\cdashline{2-3}\noalign{\vskip 0.75ex}
        & \makecell{\multicolumn{1}{p{.12\textwidth}}{\textbf{Validity}}} & \makecell{
            \multicolumn{1}{p{1.05\textwidth}}{1. The method shows a fundamental misunderstanding of the research problem and lacks any credible alignment with established scientific principles or relevant studies.} \\
            \multicolumn{1}{p{1.05\textwidth}}{2. The method partially addresses the research problem but exhibits significant flaws in its scientific underpinning, making its validity questionable despite some alignment with existing literature.} \\
            \multicolumn{1}{p{1.05\textwidth}}{3. The method adequately addresses the research problem but with some limitations in its scientific validity, showing a mix of strengths and weaknesses in its alignment with related studies.} \\
            \multicolumn{1}{p{1.05\textwidth}}{4. The method effectively addresses the research problem, demonstrating a strong scientific basis and sound alignment with existing literature, albeit with minor areas for improvement.} \\
            \multicolumn{1}{p{1.05\textwidth}}{5. The method exemplifies an exceptional understanding of the research problem, grounded in a robust scientific foundation, and shows exemplary integration and advancement of existing studies' findings.}
        } \\
        \noalign{\vskip 0.25ex}\cdashline{2-3}\noalign{\vskip 0.75ex}
        & \makecell{\multicolumn{1}{p{.12\textwidth}}{\textbf{Rigorousness}}} & \makecell{
            \multicolumn{1}{p{1.05\textwidth}}{1. The method demonstrates a fundamental lack of systematic approach, with significant inconsistencies and inaccuracies in addressing the research problem, showing a disregard for established research standards.} \\
            \multicolumn{1}{p{1.05\textwidth}}{2. The method shows a minimal level of systematic effort but is marred by notable inaccuracies, lack of precision, and inconsistencies that undermine the rigorousness of the method in tackling the research problem.} \\
            \multicolumn{1}{p{1.05\textwidth}}{3. The method exhibits an average level of systematic structure and adherence to research standards but lacks the thoroughness, precision, and consistency required for a rigorous scientific inquiry.} \\
            \multicolumn{1}{p{1.05\textwidth}}{4. The method is well-structured and systematic, with a good level of precision and consistency, indicating a strong adherence to research standards, though it falls short of exemplifying the highest level of rigorousness.} \\
            \multicolumn{1}{p{1.05\textwidth}}{5. The method exemplifies exceptional rigorousness, with outstanding thoroughness, precision, and consistency in its systematic approach, setting a benchmark for high standards in scientific research quality.}
        } \\
        \noalign{\vskip 0.25ex}\cdashline{2-3}\noalign{\vskip 0.75ex}
        & \makecell{\multicolumn{1}{p{.12\textwidth}}{\textbf{Innovativeness}}} & \makecell{
            \multicolumn{1}{p{1.05\textwidth}}{1. The method introduces no novel elements, fully relying on existing techniques without any attempt to modify or adapt them for the specific research problem, showing a lack of innovativeness.} \\
            \multicolumn{1}{p{1.05\textwidth}}{2. The method shows minimal innovation, with only slight modifications to existing techniques that do not substantially change or improve the approach to the research problem.} \\
            \multicolumn{1}{p{1.05\textwidth}}{3. The method demonstrates moderate innovativeness, incorporating known techniques with some new elements or combinations that offer a somewhat fresh approach to the research problem but fall short of a significant breakthrough.} \\
            \multicolumn{1}{p{1.05\textwidth}}{4. The method is highly innovative, introducing new techniques or novel combinations of existing methods that significantly differ from standard practices, offering a new perspective or solution to the research problem.} \\
            \multicolumn{1}{p{1.05\textwidth}}{5. The method represents a groundbreaking innovation, fundamentally transforming the approach to the research problem with novel techniques or methodologies that redefine the field's standard practices.}
        } \\
        \noalign{\vskip 0.25ex}\cdashline{2-3}\noalign{\vskip 0.75ex}
        & \makecell{\multicolumn{1}{p{.12\textwidth}}{\textbf{Generalizability}}} & \makecell{
            \multicolumn{1}{p{1.05\textwidth}}{1. The method shows no adaptability, failing to extend its applicability beyond its original context or dataset, showing a complete lack of generalizability.} \\
            \multicolumn{1}{p{1.05\textwidth}}{2. The method demonstrates minimal adaptability, with limited evidence of potential applicability to contexts slightly different from the original.} \\
            \multicolumn{1}{p{1.05\textwidth}}{3. The method exhibits some level of adaptability, suggesting it could be applicable to related contexts or datasets with modifications.} \\
            \multicolumn{1}{p{1.05\textwidth}}{4. The method is adaptable and shows evidence of applicability to a variety of contexts or datasets beyond the original.} \\
            \multicolumn{1}{p{1.05\textwidth}}{5. The method is highly adaptable, demonstrating clear evidence of broad applicability across diverse contexts, populations, and settings.}
        } \\
        \bottomrule
        \end{tabular}
    }
\end{table*}

\begin{table*}
    \caption{The criteria induced from human judgments for validating the experiment designs, which are used to align model-based evaluations with actual human preferences.}
    \label{tab:criteria:experiment}
    \vspace{-0.1in}
    \resizebox{0.95\textwidth}{!}{
    \renewcommand{\arraystretch}{1.1}
    \renewcommand{\tabcolsep}{2.5mm}
        \begin{tabular}{lll}
        \toprule
        \multicolumn{1}{p{.12\textwidth}}{\textbf{Types}} & \makecell{\multicolumn{1}{p{.12\textwidth}}{\textbf{Criteria}}} & \makecell{\multicolumn{1}{p{1.05\textwidth}}{\textbf{Texts}}} \\
        \midrule
        \multirow{50}{*}{\textbf{Experiment}} 
        & \makecell{\multicolumn{1}{p{.12\textwidth}}{\textbf{Clarity}}} & \makecell{
            \multicolumn{1}{p{1.05\textwidth}}{1. The experiment design is extremely unclear, with critical details missing or ambiguous, making it nearly impossible for others to understand the setup, procedure, or expected outcomes.} \\
            \multicolumn{1}{p{1.05\textwidth}}{2. The experiment design lacks significant clarity, with many important aspects poorly explained or omitted, challenging others to grasp the essential elements of the setup, procedure, or expected outcomes.} \\
            \multicolumn{1}{p{1.05\textwidth}}{3. The experiment design is moderately clear, but some aspects are not detailed enough, leaving room for interpretation or confusion about the setup, procedure, or expected outcomes.} \\
            \multicolumn{1}{p{1.05\textwidth}}{4. The experiment design is mostly clear, with most aspects well-described, allowing others to understand the setup, procedure, and expected outcomes with minimal ambiguity.} \\
            \multicolumn{1}{p{1.05\textwidth}}{5. The experiment design is exceptionally clear, precise, and detailed, enabling easy understanding of the setup, procedure, and expected outcomes, with no ambiguity or need for further clarification.}
        } \\
        \noalign{\vskip 0.25ex}\cdashline{2-3}\noalign{\vskip 0.75ex}
        & \makecell{\multicolumn{1}{p{.12\textwidth}}{\textbf{Validity}}} & \makecell{
            \multicolumn{1}{p{1.05\textwidth}}{1. The experiment design demonstrates a fundamental misunderstanding of the research problem, lacks alignment with scientific methods, and shows no evidence of validity in addressing the research questions or testing the proposed methods.} \\
            \multicolumn{1}{p{1.05\textwidth}}{2. The experiment design has significant flaws in its approach to the research problem and scientific method, with minimal or questionable evidence of validity, making it largely ineffective in addressing the research questions or testing the proposed methods.} \\
            \multicolumn{1}{p{1.05\textwidth}}{3. The experiment design is generally aligned with the research problem and scientific method but has some limitations in its validity, offering moderate evidence that it can somewhat effectively address the research questions or test the proposed methods.} \\
            \multicolumn{1}{p{1.05\textwidth}}{4. The experiment design is well-aligned with the research problem and scientific method, providing strong evidence of validity and effectively addressing the research questions and testing the proposed methods, despite minor limitations.} \\
            \multicolumn{1}{p{1.05\textwidth}}{5. The experiment design excellently aligns with the research problem and scientific method, demonstrating robust evidence of validity and outstandingly addressing the research questions and testing the proposed methods without significant limitations.}
        } \\
        \noalign{\vskip 0.25ex}\cdashline{2-3}\noalign{\vskip 0.75ex}
        & \makecell{\multicolumn{1}{p{.12\textwidth}}{\textbf{Robustness}}} & \makecell{
            \multicolumn{1}{p{1.05\textwidth}}{1. The experiment design demonstrates a fundamental lack of understanding of the scientific method, with no evidence of durability or adaptability across varying conditions, leading to highly unreliable and non-replicable results.} \\
            \multicolumn{1}{p{1.05\textwidth}}{2. The experiment design shows minimal consideration for robustness, with significant oversights in addressing variability and ensuring consistency across different scenarios, resulting in largely unreliable outcomes.} \\
            \multicolumn{1}{p{1.05\textwidth}}{3. The experiment design adequately addresses some aspects of robustness but lacks comprehensive measures to ensure durability and consistency across a wide range of conditions, leading to moderate reliability.} \\
            \multicolumn{1}{p{1.05\textwidth}}{4. The experiment design incorporates a solid understanding of robustness, with clear efforts to ensure the experiment's durability and consistency across diverse conditions, though minor improvements are still possible for optimal reliability.} \\
            \multicolumn{1}{p{1.05\textwidth}}{5. The experiment design exemplifies an exceptional commitment to robustness, with meticulous attention to durability and adaptability across all possible conditions, ensuring highly reliable and universally applicable results.}
        } \\
        \noalign{\vskip 0.25ex}\cdashline{2-3}\noalign{\vskip 0.75ex}
        & \makecell{\multicolumn{1}{p{.12\textwidth}}{\textbf{Feasibility}}} & \makecell{
            \multicolumn{1}{p{1.05\textwidth}}{1. The experiment design is fundamentally unfeasible, with insurmountable resource, time, or technological constraints that make implementation virtually impossible within the proposed framework.} \\
            \multicolumn{1}{p{1.05\textwidth}}{2. The experiment design faces significant feasibility challenges, including major resource, time, or technological limitations, that heavily compromise its practical execution and likelihood of success.} \\
            \multicolumn{1}{p{1.05\textwidth}}{3. The experiment design is somewhat feasible, with moderate constraints on resources, time, or technology that could be addressed with adjustments, though these may not guarantee success.} \\
            \multicolumn{1}{p{1.05\textwidth}}{4. The experiment design is largely feasible, with minor resource, time, or technological limitations that can be effectively managed or mitigated, ensuring a high probability of successful implementation.} \\
            \multicolumn{1}{p{1.05\textwidth}}{5. The experiment design is highly feasible, with no significant constraints on resources, time, or technology, indicating that it can be implemented smoothly and successfully within the proposed framework.}
        } \\
        \noalign{\vskip 0.25ex}\cdashline{2-3}\noalign{\vskip 0.75ex}
        & \makecell{\multicolumn{1}{p{.12\textwidth}}{\textbf{Reproducibility}}} & \makecell{
            \multicolumn{1}{p{1.05\textwidth}}{1. The experiment design lacks critical details, making it virtually impossible for other researchers to replicate the study under the same conditions or methodologies.} \\
            \multicolumn{1}{p{1.05\textwidth}}{2. The experiment provides some essential information but omits significant details needed for replication, leading to considerable ambiguity in methodology or conditions.} \\
            \multicolumn{1}{p{1.05\textwidth}}{3. The experiment design includes sufficient details for replication, but lacks clarity or completeness in certain areas, posing challenges for seamless reproducibility.} \\
            \multicolumn{1}{p{1.05\textwidth}}{4. The experiment is well-documented with clear, detailed instructions and methodologies that allow for consistent replication, albeit with minor areas for improvement.} \\
            \multicolumn{1}{p{1.05\textwidth}}{5. The experiment design is exemplary in its clarity, detail, and comprehensiveness, ensuring that other researchers can precisely and effortlessly replicate the study under identical conditions and methodologies.}
        } \\
        \bottomrule
        \end{tabular}
    }
\end{table*}
\onecolumn

\begin{center}

\begingroup

\fontsize{6.5pt}{9pt}\selectfont

\begin{longtable}{p{.05\textwidth}p{.08\textwidth}p{.75\textwidth}}
\caption{The examples of research idea generation results from the proposed full ResearchAgent.} 
\vspace{-0.1in}
\label{tab:examples} \\

\hline 
\textbf{Index} & \textbf{Types} & \hspace{0.05in} \textbf{Texts}
\\ \hline 
\endfirsthead

\multicolumn{3}{c}%
{{\bfseries \tablename\ \thetable{} -- Continued from the previous page}} \\
\hline 
\textbf{Index} & \textbf{Types} & \hspace{0.05in} \textbf{Texts}
\\ \hline 
\endhead 

\multicolumn{3}{r}{\textit{\textbf{Continued on the next page}}} \\ \hline
\endfoot

\hline \hline
\endlastfoot

%%%%%%%%%%%%%%%%%%%%%%%%%%%%%% Examples %%%%%%%%%%%%%%%%%%%
\multirow{4}{*}{\textbf{1}} & 
\multicolumn{1}{p{.08\textwidth}}{\textbf{Input}} & 
\makecell{
    \multicolumn{1}{p{.75\textwidth}}{\textbf{Title}: Knowledge-Augmented Language Model Prompting for Zero-Shot Knowledge Graph Question Answering} \\
    \noalign{\vskip 0.25ex}\cdashline{0-0}\noalign{\vskip 0.75ex}
    \multicolumn{1}{p{.75\textwidth}}{\textbf{Abstract}: Large Language Models (LLMs) are capable of performing zero-shot closed-book question answering tasks, based on their internal knowledge stored in parameters during pre-training. However, such internalized knowledge might be insufficient and incorrect, which could lead LLMs to generate factually wrong answers. Furthermore, fine-tuning LLMs to update their knowledge is expensive. To this end, we propose to augment the knowledge directly in the input of LLMs. Specifically, we first retrieve the relevant facts to the input question from the knowledge graph based on semantic similarities between the question and its associated facts. After that, we prepend the retrieved facts to the input question in the form of the prompt, which is then forwarded to LLMs to generate the answer. Our framework, Knowledge-Augmented language model PromptING (KAPING), requires no model training, thus completely zero-shot. We validate the performance of our KAPING framework on the knowledge graph question answering task, that aims to answer the user’s question based on facts over a knowledge graph, on which ours outperforms relevant zero-shot baselines by up to 48\% in average, across multiple LLMs of various sizes.} \\
    \noalign{\vskip 0.25ex}\cdashline{0-0}\noalign{\vskip 0.75ex}
    \multicolumn{1}{p{.75\textwidth}}{\textbf{Entities}: ['Natural language', 'Learning management system', 'Question answering', 'Wikipedia', 'Artificial intelligence', 'Named-entity recognition', 'Quality assurance', 'Long short-term memory', 'English language', 'Google Books', 'Natural-language programming', 'LMS color space', 'GitHub', 'Logic learning machine', 'Machine translation', 'DBpedia', 'Integrated library system', 'Spanish language', 'Neural machine translation', 'ArXiv', 'Artificial general intelligence', 'Natural-language understanding', 'Nonlinear programming', 'Codex', 'LoRa', 'Adhesive', 'Common Crawl', 'ROUGE (metric)', 'Transformer', 'Tomas Mikolov']} \\
} \\
\cline{2-3}
& \multicolumn{1}{p{.08\textwidth}}{\textbf{Problem}} & 
\makecell{
    \multicolumn{1}{p{.75\textwidth}}{Developing a Multimodal Knowledge-Aware Prompting System for Multilingual Zero-Shot Question Answering across Structured and Unstructured Data Sources} 
} \\
\cline{2-3}
& \multicolumn{1}{p{.08\textwidth}}{\textbf{Method}} &
\makecell{
    \multicolumn{1}{p{.75\textwidth}}{The further refined and enhanced method for developing a multimodal knowledge-aware prompting system for multilingual zero-shot question answering (MMKP-ZSQA) incorporates advanced refinements in the following components to address the requirements for innovativeness, clarity, and generalizability:} \\ \\ 
    \multicolumn{1}{p{.75\textwidth}}{1. Interconnected Multimodal Knowledge Orchestration (IMKO): } \\ 
    \multicolumn{1}{p{.75\textwidth}}{- IMKO is an advanced evolution of SMKI, incorporating state-of-the-art language and vision models to handle data verbalization with context-aware disambiguation methods.} \\
    \multicolumn{1}{p{.75\textwidth}}{- Techniques like attention mechanisms are employed to weigh the relevance of different data pieces when integrated, revealing how textual and visual data augment each other in a synthesized representation.} \\ \\
    \multicolumn{1}{p{.75\textwidth}}{2. Linguistically Inclusive Retrieval Engine (LIRE): } \\ 
    \multicolumn{1}{p{.75\textwidth}}{- LIRE extends EMKA with an emphasis on semantic understanding, using transformer-based models trained on diverse datasets, including idiomatic and cultural nuances across languages.} \\
    \multicolumn{1}{p{.75\textwidth}}{- Specific algorithms to handle linguistic phenomena such as code-switching and transliteration are included, enhancing the application to a broader set of languages and dialects.} \\ \\
    \multicolumn{1}{p{.75\textwidth}}{3. Prompt Learning and Optimization Nexus (PLON): } \\ 
    \multicolumn{1}{p{.75\textwidth}}{- Building on AMPL, PLON focuses on creating a library of optimized prompts categorized by linguistic features and data modalities, using Bayesian optimization algorithms.} \\
    \multicolumn{1}{p{.75\textwidth}}{- It includes domain adaptation techniques and a wider array of meta-learning strategies with case studies for high and low-resource languages, and outlines theoretical frameworks for their implementation.} \\ \\ 
    \multicolumn{1}{p{.75\textwidth}}{4. Cross-Modal Integrative Learning System (C-MILS): } \\
    \multicolumn{1}{p{.75\textwidth}}{- C-MILS advances SCAS by detailing the use of multi-head attention across modalities for effective and scalable reasoning, with exemplifications on how each modality can enhance comprehension synergistically.} \\
    \multicolumn{1}{p{.75\textwidth}}{- The component now incorporates a layer of abstraction to distill knowledge into a modality-agnostic format, aiding reasoning and facilitating interpretability across languages and data types.} \\ \\ 
    \multicolumn{1}{p{.75\textwidth}}{5. User-Centric Adaptation and Privacy Framework (UCAPF): } \\
    \multicolumn{1}{p{.75\textwidth}}{- UCAPF enriches ALCUM with a user-centric design, focusing on interaction protocols that describe user engagement cycles, feedback loops, and privacy-preserving active learning paradigms, with practical workflows and pseudocode.} \\
    \multicolumn{1}{p{.75\textwidth}}{- Aligns with GDPR and other privacy frameworks to fortify trust and address the ethical use of data in dynamic learning environments.} \\ \\ 
    \multicolumn{1}{p{.75\textwidth}}{6. Global Evaluation and Reflective Testing System (GERTS): } \\
    \multicolumn{1}{p{.75\textwidth}}{- GERTS expands CEB by presenting a multi-tiered validation and reflection methodology to adjust system components based on a diverse set of metrics, including fairness, bias, interpretability, and computational efficiency.} \\
    \multicolumn{1}{p{.75\textwidth}}{- Offers a structured approach to cross-cultural evaluation, including the use of demographic and regional diversity in forming test cohorts.} \\ 
} \\
\cline{2-3}
& \multicolumn{1}{p{.08\textwidth}}{\textbf{Experiment}} & 
\makecell{
    \multicolumn{1}{p{.75\textwidth}}{The experiment, named "Refined Experiment for Multimodal Knowledge-Aware Prompting System for Multilingual Zero-shot Question Answering (RE-MKP-ZSQA)", aims to methodically develop and validate an advanced AI system. The experiment is streamlined to address feasibility, clarity, and reproducibility concerns while upholding robustness and validity by adhering to the following refined phases:} \\ \\ 
    \multicolumn{1}{p{.75\textwidth}}{1. Detailed System Implementation Plan:} \\
    \multicolumn{1}{p{.75\textwidth}}{- Provide a publicly accessible project roadmap with specific milestones, resource allocation, and timelines.} \\ \\
    \multicolumn{1}{p{.75\textwidth}}{2. Dataset Curation with Clear Guidelines:} \\
    \multicolumn{1}{p{.75\textwidth}}{- Publish precise annotation guidelines with strategies to prevent bias.} \\
    \multicolumn{1}{p{.75\textwidth}}{- Document the dataset assembly process, including source selection and data processing procedures.} \\ \\
    \multicolumn{1}{p{.75\textwidth}}{3. Transparent System Training:} \\
    \multicolumn{1}{p{.75\textwidth}}{- Offer a detailed training protocol with hyperparameters, optimization strategies, and Bayesian optimization processes used in PLON.} \\ \\
    \multicolumn{1}{p{.75\textwidth}}{4. Structured Zero-Shot Evaluation:} \\
    \multicolumn{1}{p{.75\textwidth}}{- Outline evaluation metrics derived from GERTS with benchmark datasets to test zero-shot capabilities.} \\ \\
    \multicolumn{1}{p{.75\textwidth}}{5. Clearer Interdisciplinary Evaluation Protocol:} \\
    \multicolumn{1}{p{.75\textwidth}}{- Specify the composition of the evaluation committee, criteria for assessments, and methods for integrating the feedback.} \\ \\
    \multicolumn{1}{p{.75\textwidth}}{6. Iterative Improvement with Validation Metrics:} \\
    \multicolumn{1}{p{.75\textwidth}}{- Describe statistical methods for reflective assessment and continuous improvement, aligned with multi-tiered GERTS methodology.} \\ \\
    \multicolumn{1}{p{.75\textwidth}}{7. User-Centric Design and Privacy Compliance Evaluation:} \\
    \multicolumn{1}{p{.75\textwidth}}{- Structure user studies with targeted data points to assess usability and cultural adaptability.} \\
    \multicolumn{1}{p{.75\textwidth}}{- Outline privacy compliance protocols to adhere to international standards.} \\ \\
    \multicolumn{1}{p{.75\textwidth}}{8. Detailed Global Scalability Evaluation Method:} \\
    \multicolumn{1}{p{.75\textwidth}}{- Define evaluation metrics for scalability tests and describe diverse infrastructural setups.} \\ \\
    \multicolumn{1}{p{.75\textwidth}}{9. Enhanced Reporting for Reproducibility:} \\
    \multicolumn{1}{p{.75\textwidth}}{- Commit to creating a comprehensive report with precise specifications, configurations, and instructions for replication purposes.} \\
    \multicolumn{1}{p{.75\textwidth}}{- Utilize GitHub for version-controlled deposition of code and datasets, and arXiv for openly accessible experiment protocols and findings.} \\
}
\\

\midrule

\multirow{25}{*}{\textbf{2}} & 
\multicolumn{1}{p{.08\textwidth}}{\textbf{Input}} & 
\makecell{
    \multicolumn{1}{p{.75\textwidth}}{\textbf{Title}: Test-Time Self-Adaptive Small Language Models for Question Answering} \\
    \noalign{\vskip 0.25ex}\cdashline{0-0}\noalign{\vskip 0.75ex}
    \multicolumn{1}{p{.75\textwidth}}{\textbf{Abstract}: Recent instruction-finetuned large language models (LMs) have achieved notable performances in various tasks, such as question-answering (QA). However, despite their ability to memorize a vast amount of general knowledge across diverse tasks, they might be suboptimal on specific tasks due to their limited capacity to transfer and adapt knowledge to target tasks. Moreover, further finetuning LMs with labeled datasets is often infeasible due to their absence, but it is also questionable if we can transfer smaller LMs having limited knowledge only with unlabeled test data. In this work, we show and investigate the capabilities of smaller self-adaptive LMs, only with unlabeled test data. In particular, we first stochastically generate multiple answers, and then ensemble them while filtering out low-quality samples to mitigate noise from inaccurate labels. Our proposed self-adaption strategy demonstrates significant performance improvements on benchmark QA datasets with higher robustness across diverse prompts, enabling LMs to stay stable.} \\
    \noalign{\vskip 0.25ex}\cdashline{0-0}\noalign{\vskip 0.75ex}
    \multicolumn{1}{p{.75\textwidth}}{\textbf{Entities}: ['Codex', 'Natural language', 'English language', 'United States', 'Question answering', 'Natural-language programming', 'GTRI Information and Communications Laboratory', 'Artificial intelligence', 'LoRa', 'Llama', 'Python (programming language)', 'Learning management system', 'Natural language processing', 'Reinforcement learning', 'LMS color space', 'Wikipedia', 'GitHub', 'Natural-language understanding', 'London, Midland and Scottish Railway', 'Integrated library system', 'Language model', 'Chinese language', 'Lumen (unit)', 'Spanish language', 'English Wikipedia', 'Logic learning machine', 'Gradient descent', 'Alternative public offering', 'Technology transfer', 'Dialogue system']} \\
} \\
\cline{2-3}
& \multicolumn{1}{p{.08\textwidth}}{\textbf{Problem}} & 
\makecell{
    \multicolumn{1}{p{.75\textwidth}}{Developing a Scalable, Domain-Adaptive Test-Time Training Protocol for Low-Resource Language QA Using Small Language Models} \\ 
} \\
\cline{2-3}
& \multicolumn{1}{p{.08\textwidth}}{\textbf{Method}} &
\makecell{
    \multicolumn{1}{p{.75\textwidth}}{1. Selection of Scalable Compact Language Models (CLMs): Identify and evaluate existing CLMs suitable for adaptation, emphasizing models with minimal computational requirements.} \\ \\
    \multicolumn{1}{p{.75\textwidth}}{2. Creation of a Multilingual Test-Time Training (TTT) Framework: Develop a TTT protocol that enables CLMs to adapt to new domains and languages during the inference phase, leveraging unsupervised learning techniques and pseudo-label generation.} \\ \\
    \multicolumn{1}{p{.75\textwidth}}{3. Synthetic and Unsupervised Data Generation: Utilize a combination of unsupervised and synthetic data generation methods to produce multilingual QA pairs, employing techniques such as back-translation and context-based question synthesis.} \\ \\
    \multicolumn{1}{p{.75\textwidth}}{4. Domain-Adaptive Mechanisms: Introduce domain-adaptive components, including feature adaptation layers and meta-learning algorithms, which tailor the model's behavior to new contexts and languages at test time.} \\ \\
    \multicolumn{1}{p{.75\textwidth}}{5. Incremental Language Addition and Dominance Assessment: Start with a subset of linguistically diverse, low-resource languages. Evaluate domain adaptability for each language via an iterative process, ensuring models learn to prioritize resource efficiency.} \\ \\
    \multicolumn{1}{p{.75\textwidth}}{6. Model Robustness and Generalization: Perform robustness tuning (RT) to prepare models for unforeseen linguistic variations and conduct thorough evaluations across multiple domains to ensure models can generalize their learning effectively.} \\ \\
    \multicolumn{1}{p{.75\textwidth}}{7. Human-In-The-Loop Evaluation: Conduct evaluations with native speakers and domain experts to validate the relevance and accuracy of the QA outputs, incorporating feedback into the iterative training process.} \\ \\
    \multicolumn{1}{p{.75\textwidth}}{8. Open-Sourcing and Community Collaboration: Make the TTT protocol, trained models, and evaluation benchmarks publicly available for the research community, fostering collaboration and further innovation.} \\ 
} \\
\cline{2-3}
& \multicolumn{1}{p{.08\textwidth}}{\textbf{Experiment}} & 
\makecell{ 
    \multicolumn{1}{p{.75\textwidth}}{1. Selection and Preparation:} \\
    \multicolumn{1}{p{.75\textwidth}}{- Identify potential compact language models (CLMs) suitable for domain adaptation and test-time training, focusing on those with minimal computational requirement and the ability to be fine-tuned or adapted in an unsupervised manner.} \\
    \multicolumn{1}{p{.75\textwidth}}{- Prepare a diverse set of low-resource languages and corresponding text corpora, ensuring linguistic diversity and sociocultural significance. Select benchmark datasets for these languages if available.} \\ \\
    \multicolumn{1}{p{.75\textwidth}}{2. Training and Adaptation Procedure:} \\
    \multicolumn{1}{p{.75\textwidth}}{- Create a Test-Time Training (TTT) framework that allows selected CLMs to adapt to various domains in the selected low-resource languages during the inference phase.} \\
    \multicolumn{1}{p{.75\textwidth}}{- Implement unsupervised learning techniques and pseudo-label generation to produce QA pairs, utilizing back-translation and context-based question synthesis to generate synthetic datasets for languages with limited or no available QA datasets.} \\
    \multicolumn{1}{p{.75\textwidth}}{- Integrate domain-adaptive components and meta-learning algorithms into the CLMs to enable domain-specific adaptations at test time.
} \\ \\
    \multicolumn{1}{p{.75\textwidth}}{3. Iterative Evaluation and Refinement:} \\
    \multicolumn{1}{p{.75\textwidth}}{- Begin adaptation and training with a single low-resource language and gradually add additional languages, monitoring the domain adaptability and model performance metrics after each addition.} \\ 
    \multicolumn{1}{p{.75\textwidth}}{- Perform robustness tuning and cross-domain evaluations for each CLM and language adaptation to ensure generalizability and prevent overfitting.} \\ \\
    \multicolumn{1}{p{.75\textwidth}}{4. Human-In-The-Loop Assessment:} \\
    \multicolumn{1}{p{.75\textwidth}}{- Enlist native speakers and domain experts to evaluate the relevance and accuracy of the model's QA outputs for each language.} \\ 
    \multicolumn{1}{p{.75\textwidth}}{- Incorporate feedback into the iterative training process, refining and re-adapting the models accordingly.} \\ \\
    \multicolumn{1}{p{.75\textwidth}}{5. Open-Sourcing and Community Feedback:} \\
    \multicolumn{1}{p{.75\textwidth}}{- Make the TTT protocol, adaptive CLMs, evaluation benchmarks, and any synthetic datasets publicly available for the research community.} \\ \\
    \multicolumn{1}{p{.75\textwidth}}{6. Experiment Monitoring and Documentation:} \\
    \multicolumn{1}{p{.75\textwidth}}{- Record all the parameters, datasets, model configurations, and evaluation metrics meticulously to ensure robustness and reproducibility.} \\ 
    \multicolumn{1}{p{.75\textwidth}}{- Document any challenges faced, unexpected results, or adaptions made during the experiment for open-sourcing purposes.} \\ \\
    \multicolumn{1}{p{.75\textwidth}}{7. Data Analysis and Reporting:} \\
    \multicolumn{1}{p{.75\textwidth}}{- Analyze the collected performance data quantitatively, using appropriate statistical methods to compare with non-adaptive baselines.} \\
    \multicolumn{1}{p{.75\textwidth}}{- Report qualitative findings from human-in-the-loop evaluations, interpreting the implications for language model performance in low-resource language domains.}
}
\\

\midrule

\multirow{2}{*}{\textbf{3}} & 
\multicolumn{1}{p{.08\textwidth}}{\textbf{Input}} & 
\makecell{
    \multicolumn{1}{p{.75\textwidth}}{\textbf{Title}: Whole-brain annotation and multi-connectome cell typing quantifies circuit stereotypy in Drosophila} \\
    \noalign{\vskip 0.25ex}\cdashline{0-0}\noalign{\vskip 0.75ex}
    \multicolumn{1}{p{.75\textwidth}}{\textbf{Abstract}: The fruit fly Drosophila melanogaster combines surprisingly sophisticated behaviour with a highly tractable nervous system. A large part of the fly’s success as a model organism in modern neuroscience stems from the concentration of collaboratively generated molecular genetic and digital resources. As presented in our FlyWire companion paper1, this now includes the first full brain connectome of an adult animal. Here we report the systematic and hierarchical annotation of this 130,000-neuron connectome including neuronal classes, cell types and developmental units (hemilineages). This enables any researcher to navigate this huge dataset and find systems and neurons of interest, linked to the literature through the Virtual Fly Brain database2. Crucially, this resource includes 4,552 cell types. 3,094 are rigorous consensus validations of cell types previously proposed in the “hemibrain” connectome3. In addition, we propose 1,458 new cell types, arising mostly from the fact that the FlyWire connectome spans the whole brain, whereas the hemibrain derives from a subvolume. Comparison of FlyWire and the hemibrain showed that cell type counts and strong connections were largely stable, but connection weights were surprisingly variable within and across animals. Further analysis defined simple heuristics for connectome interpretation: connections stronger than 10 unitary synapses or providing >1\% of the input to a target cell are highly conserved. Some cell types showed increased variability across connectomes: the most common cell type in the mushroom body, required for learning and memory, is almost twice as numerous in FlyWire as the hemibrain. We find evidence for functional homeostasis through adjustments of the absolute amount of excitatory input while maintaining the excitation-inhibition ratio. Finally, and surprisingly, about one third of the cell types proposed in the hemibrain connectome could not yet be reliably identified in the FlyWire connectome. We therefore suggest that cell types should be defined to be robust to inter-individual variation, namely as groups of cells that are quantitatively more similar to cells in a different brain than to any other cell in the same brain. Joint analysis of the FlyWire and hemibrain connectomes demonstrates the viability and utility of this new definition. Our work defines a consensus cell type atlas for the fly brain and provides both an intellectual framework and open source toolchain for brain-scale comparative connectomics.} \\
    \noalign{\vskip 0.25ex}\cdashline{0-0}\noalign{\vskip 0.75ex}
    \multicolumn{1}{p{.75\textwidth}}{\textbf{Entities}: ['Virtual Fly Brain', 'Central nervous system', 'Transposable element', 'SUMO protein', 'Kenyon cell', 'Romani people', 'Induced stem cells', 'Ventral nerve cord', 'FlyBase', "Parkinson's disease", 'Virtual Network Computing', 'P element', 'Piwi-interacting RNA', 'Drosophila Genetic Reference Panel', 'Bateson–Dobzhansky–Muller model', 'J. B. S. Haldane', 'ATG7', "Haldane's rule", 'Oxford Nanopore Technologies', 'Drosophila mauritiana', 'Germline', 'PINK1', 'Migratory locust', 'CRISPR', 'Heliconius', 'GINS (protein complex)', 'Parkin (ligase)', 'Lepidoptera', 'Illumina, Inc.', 'Drosophila']} \\
} \\
\cline{2-3}
& \multicolumn{1}{p{.08\textwidth}}{\textbf{Problem}} & 
\makecell{
    \multicolumn{1}{p{.75\textwidth}}{Investigating the Functional Implications of Connectome Variability in Drosophila's Learning and Memory Circuits Across Different Environmental and Genetic Contexts} \\ 
} \\
\cline{2-3}
& \multicolumn{1}{p{.08\textwidth}}{\textbf{Method}} &
\makecell{
    \multicolumn{1}{p{.75\textwidth}}{The proposed method involves a multi-tiered approach that integrates connectomics, behavioral assays, genetic manipulation, and computational modeling to investigate the functional implications of connectome variability in Drosophila's learning and memory circuits. The method consists of the following steps:} \\ \\
    \multicolumn{1}{p{.75\textwidth}}{1. Connectome Mapping and Variability Analysis:} \\ 
    \multicolumn{1}{p{.75\textwidth}}{a. Utilize the Virtual Fly Brain database to identify and compare individual connectomes, focusing on the mushroom body.} \\ 
    \multicolumn{1}{p{.75\textwidth}}{b. Quantify the variability in connection weights and cell type counts using statistical methods and machine learning algorithms to identify patterns of variability.} \\ \\
    \multicolumn{1}{p{.75\textwidth}}{2. Behavioral Assays:} \\
    \multicolumn{1}{p{.75\textwidth}}{a. Design a series of learning and memory tasks for Drosophila, such as olfactory conditioning or visual pattern recognition.} \\
    \multicolumn{1}{p{.75\textwidth}}{b. Test groups of flies with known connectome profiles under controlled environmental conditions to establish baseline behavioral data.} \\ \\
    \multicolumn{1}{p{.75\textwidth}}{3. Environmental and Genetic Perturbations:} \\ 
    \multicolumn{1}{p{.75\textwidth}}{a. Expose different groups of flies to varied learning paradigms and sensory inputs to create environmental perturbations.} \\ 
    \multicolumn{1}{p{.75\textwidth}}{b. Use CRISPR-Cas9 technology to introduce targeted mutations in genes like PINK1 or Parkin, creating genetic perturbations.} \\ 
    \multicolumn{1}{p{.75\textwidth}}{c. Assess the impact of these perturbations on connectome structure using high-resolution imaging and reconstruction techniques.} \\ \\
    \multicolumn{1}{p{.75\textwidth}}{4. Transcriptomic and Spatial Analysis:} \\ 
    \multicolumn{1}{p{.75\textwidth}}{a. Apply single-cell RNA sequencing and spatial transcriptomics to profile gene expression changes in response to environmental and genetic perturbations.} \\ 
    \multicolumn{1}{p{.75\textwidth}}{b. Correlate transcriptomic data with connectome changes to identify molecular pathways associated with structural and functional plasticity.} \\ \\
    \multicolumn{1}{p{.75\textwidth}}{5. Computational Modeling and Network Analysis:} \\ 
    \multicolumn{1}{p{.75\textwidth}}{a. Develop computational models to simulate the effects of connectome variability on neural circuit function.} \\ 
    \multicolumn{1}{p{.75\textwidth}}{b. Use network analysis tools to explore information flow and circuit dynamics, integrating data from related papers on olfactory projection neurons and information flow in the olfactory system.} \\ \\
    \multicolumn{1}{p{.75\textwidth}}{6. Synthesis and Validation:} \\ 
    \multicolumn{1}{p{.75\textwidth}}{a. Integrate findings from behavioral assays, connectome mapping, transcriptomic analysis, and computational modeling.} \\ 
    \multicolumn{1}{p{.75\textwidth}}{b. Validate the proposed models and hypotheses through iterative experimentation and refinement.} \\ 
} \\
\cline{2-3}
& \multicolumn{1}{p{.08\textwidth}}{\textbf{Experiment}} & 
\makecell{ 
    \multicolumn{1}{p{.75\textwidth}}{1. Pre-experimental Setup:} \\
    \multicolumn{1}{p{.75\textwidth}}{a. Establish a Drosophila melanogaster breeding program to ensure a consistent supply of genetically similar flies for the experiments.} \\
    \multicolumn{1}{p{.75\textwidth}}{b. Develop a standardized protocol for rearing flies under controlled conditions to minimize pre-experimental variability.} \\ 
    \multicolumn{1}{p{.75\textwidth}}{c. Select and validate CRISPR-Cas9 constructs for targeted gene editing of PINK1, Parkin, and other genes of interest.} \\ 
    \multicolumn{1}{p{.75\textwidth}}{d. Train personnel in the use of the Virtual Fly Brain database and relevant computational tools for connectome analysis.} \\ \\
    \multicolumn{1}{p{.75\textwidth}}{2. Connectome Mapping and Variability Analysis:} \\
    \multicolumn{1}{p{.75\textwidth}}{a. Randomly assign individual flies to either a control group or various treatment groups (environmental and genetic perturbations).} \\
    \multicolumn{1}{p{.75\textwidth}}{b. Utilize high-resolution imaging techniques to map the connectomes of flies from each group, with a focus on the mushroom body.} \\
    \multicolumn{1}{p{.75\textwidth}}{c. Apply statistical and machine learning algorithms to quantify and compare the variability in connection weights and cell type counts across groups.} \\ \\
    \multicolumn{1}{p{.75\textwidth}}{3. Behavioral Assays:} \\
    \multicolumn{1}{p{.75\textwidth}}{a. Design and validate a series of learning and memory tasks, such as olfactory conditioning and visual pattern recognition, ensuring tasks are sensitive to subtle differences in performance.} \\ 
    \multicolumn{1}{p{.75\textwidth}}{b. Test flies from each group in the behavioral tasks and record performance metrics.} \\ 
    \multicolumn{1}{p{.75\textwidth}}{c. Analyze behavioral data to establish correlations with connectome profiles.} \\ \\
    \multicolumn{1}{p{.75\textwidth}}{4. Environmental and Genetic Perturbations:} \\
    \multicolumn{1}{p{.75\textwidth}}{a. Expose flies to different learning paradigms and sensory inputs to induce environmental perturbations.} \\ 
    \multicolumn{1}{p{.75\textwidth}}{b. Perform gene editing using CRISPR-Cas9 to create genetic perturbations in the treatment groups.} \\ 
    \multicolumn{1}{p{.75\textwidth}}{c. Re-map connectomes post-perturbation to assess structural changes.} \\ \\
    \multicolumn{1}{p{.75\textwidth}}{5. Transcriptomic and Spatial Analysis:} \\
    \multicolumn{1}{p{.75\textwidth}}{a. Collect brain tissue from flies post-behavioral assays and perform single-cell RNA sequencing and spatial transcriptomics.} \\ 
    \multicolumn{1}{p{.75\textwidth}}{b. Analyze transcriptomic data to identify gene expression changes and correlate these with observed connectome and behavioral variations.} \\ \\
    \multicolumn{1}{p{.75\textwidth}}{6. Computational Modeling and Network Analysis:} \\
    \multicolumn{1}{p{.75\textwidth}}{a. Develop computational models to simulate the impact of observed connectome variability on neural circuit function.} \\ 
    \multicolumn{1}{p{.75\textwidth}}{b. Use network analysis to integrate behavioral, connectomic, and transcriptomic data, focusing on information flow and circuit dynamics.} \\ \\
    \multicolumn{1}{p{.75\textwidth}}{7. Synthesis and Validation:} \\
    \multicolumn{1}{p{.75\textwidth}}{a. Integrate findings across all experimental components to formulate a cohesive understanding of the functional implications of connectome variability.} \\
    \multicolumn{1}{p{.75\textwidth}}{b. Validate models and refine hypotheses through additional targeted experiments, informed by initial findings.}
}
\\

%%%%%%%%%%%%%%%%%%%%%%%%%%%%%%%%%%%%%%%%%%%%%%%%%%%%%%%%%%

\end{longtable}

\endgroup

\end{center}

\twocolumn

\end{document}